\documentclass[10pt,twocolumn,letterpaper]{article}

\usepackage{wacv}
\usepackage{times}
\usepackage{epsfig}
\usepackage{graphicx}
\usepackage{amsmath}
\usepackage{amssymb}
\usepackage{multirow}
\usepackage{xcolor}
\usepackage{adjustbox}

\usepackage[toc,page]{appendix}
\usepackage{caption}
\usepackage{comment}
\usepackage{subcaption}
\usepackage{enumitem}
\usepackage[ruled,vlined]{algorithm2e}
\usepackage[accsupp]{axessibility}  

\usepackage[pagebackref=true,breaklinks=true,letterpaper=true,colorlinks,bookmarks=false]{hyperref}

\wacvfinalcopy 

\def\wacvPaperID{510} 

\begin{document}

\title{Adversarial Open Domain Adaptation for Sketch-to-Photo Synthesis}

\author{Xiaoyu Xiang$^1$\thanks{The author was with Purdue University when conducting the work in this paper during an internship at ByteDance. She is now with Facebook.},  Ding Liu$^2$,  Xiao Yang$^2$, Yiheng Zhu$^2$, Xiaohui Shen$^2$, Jan P. Allebach$^1$ \\
$^{1}$Purdue University, $^{2}$ByteDance Inc. \\
 \tt\small \{xiang43,allebach\}@purdue.edu, \\
 \tt\small{\{liuding,yangxiao.0,yiheng.zhu,shenxiaohui\}@bytedance.com}
}

\maketitle

\begin{abstract}
   In this paper, we explore open-domain sketch-to-photo translation, which aims to synthesize a realistic photo from a freehand sketch with its class label, even if the sketches of that class are missing in the training data. It is challenging due to the lack of training supervision and the large geometric distortion between the freehand sketch and photo domains. To synthesize the absent freehand sketches from photos, we propose a framework that jointly learns sketch-to-photo and photo-to-sketch generation. However, the generator trained from fake sketches might lead to unsatisfying results when dealing with sketches of missing classes, due to the domain gap between synthesized sketches and real ones. To alleviate this issue, we further propose a simple yet effective open-domain sampling and optimization strategy to ``fool" the generator into treating fake sketches as real ones. Our method takes advantage of the learned sketch-to-photo and photo-to-sketch mapping of in-domain data and generalizes it to the open-domain classes. We validate our method on the Scribble and SketchyCOCO datasets. Compared with the recent competing methods, our approach shows impressive results in synthesizing realistic color, texture, and maintaining the geometric composition for various categories of open-domain sketches.
\end{abstract}


\section{Introduction}
Freehand sketch is an intuitive way for users to interact on visual media and express their intentions. The popularization of touch screens provides more and more scenarios for sketch-based application, \eg sketch-based photo-editing~\cite{portenier2018faceshop,dekel2018sparse,jo2019sc,olszewski2020intuitive,yang2020deep}, sketch-based image retrieval for 2D images~\cite{yu2016sketch,song2017deep,liu2017deep,zhang2018generative,yelamarthi2018zero,pang2019generalising,dey2019doodle,collomosse2019livesketch,dutta2019semantically,bhunia2020sketch,liu2020scenesketcher, Bhunia_2021_CVPR_2} and 3D shapes~\cite{wang2015sketch,zhu2016learning,dai2017deep,xie2017learning,chen2018deep}, and 3D modeling from sketches~\cite{lun20173d,han2017deepsketch2face,shen2019deepsketchhair}.

\begin{figure}[t]
    \centering
    \includegraphics[width=1.0\linewidth]{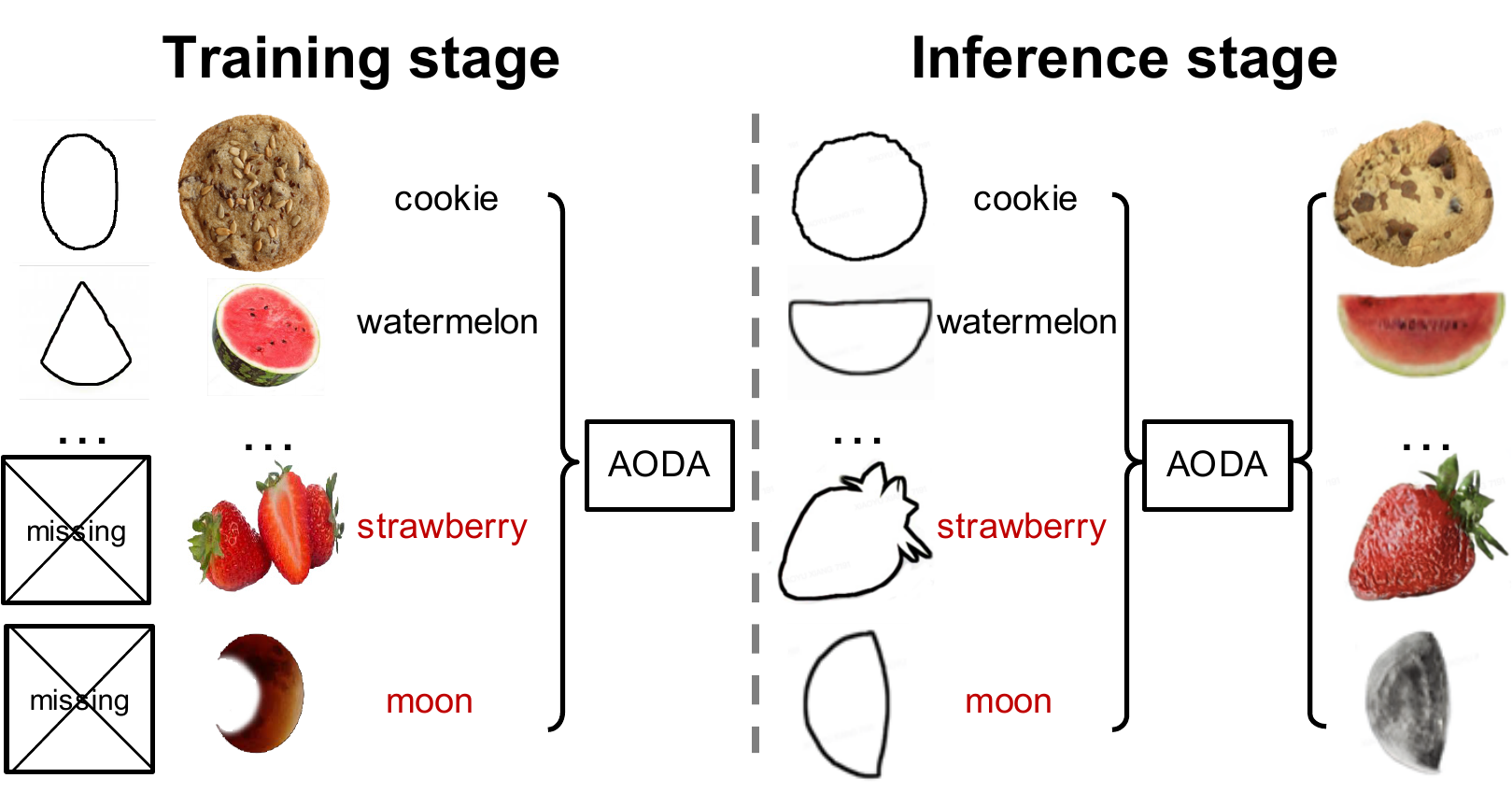}
    \caption{Illustration of open-domain sketch-to-photo synthesis problem. During the training stage of multi-class sketch-to-photo generation, sketches of some categories are missing. 
    In the inference stage, our algorithm synthesizes photos from the input sketches for not only known classes, but also the classes that were missing during the training. }
    \label{fig:illustration}
\end{figure}

Sketch-to-photo translation aims to automatically translate a sketch in the source domain $S$ to the target photo-realistic domain $P$. Many existing works~\cite{isola2017image,chen2018sketchygan,lu2018image,ghosh2019interactive,li2019linestofacephoto,gao2020sketchycoco,chen2020deepfacedrawing,li2020deepfacepencil} adopt generative adversarial networks (GAN)~\cite{goodfellow2014generative} to learn the sketch-to-image process from paired data. However, the sketch-to-photo translation task suffers from the open-domain adaptation problem, where the majority of data is unlabeled and unpaired~\cite{eitz2012humans,yi2014bmvc,li2017deeper,ha2017neural,zou2018sketchyscene,caesar2018coco,li2019photo}, and the freehand sketch covers only a small portion of the photo categories~\cite{sangkloy2016sketchy,yu2016sketch,song2017deep,liu2019unpaired,ghosh2019interactive} due to the fact that they require a large number of human annotations. Therefore, some works~\cite{isola2017image,li2019linestofacephoto,chen2020deepfacedrawing,li2020deepfacepencil} use edges extracted from the target photos as substitution. Still, edges and freehand sketches are very different: freehand sketches are human abstractions of an object, usually with more deformations. Due to this domain gap, models trained on the edge inputs easily fail to generalize to freehand sketches. A good sketch-based image generator should not only fill the correct textures within the lines, but also correct the object structure conditioned on the input composition. 

Well-labeled freehand sketches and photos can help the translation model better understand the geometry correspondence. In recent years, ~\cite{zhu2017unpaired,liu2017unsupervised,huang2018multimodal,lee2018diverse,kim2019u,liu2019unpaired} aim to learn from unpaired sketches and photos collected separately. Even so, the existing sketch datasets cannot cover all types of photos in the open domain~\cite{panareda2017open}: the largest sketch dataset \textit{Quick, Draw!}~\cite{ha2017neural} has 345 categories, while the full ImageNet~\cite{imagenet_cvpr09} has as many as 21,841 class labels. Therefore, most categories even lack corresponding freehand sketches to train a sketch-to-image translation model.

To resolve this challenging task, we propose an Adversarial Open Domain Adaptation (AODA) framework that for the first time learns to synthesize the absent freehand sketches and makes the unsupervised open-domain adaptation possible, as illustrated in Figure~\ref{fig:illustration}. We propose to jointly learn a sketch-to-photo translation network and a photo-to-sketch translation network for mapping the open-domain photos into the sketches with the GAN priors. With the bridge of the photo-to-sketch generation, we can generalize the learned correspondence between in-domain freehand sketches and photos to open-domain categories. Still, there is an unignorable domain gap between synthesized sketches and real ones, which prevents the generator from generalizing the learned correspondence to real sketches and synthesizing realistic photos for open-domain classes. 

To further mitigate its influence on the generator and leverage the output quality of open-domain translation, we introduce a simple yet effective random-mixed sampling strategy that considers a certain proportion of fake sketches as real ones blindly for all categories. With the proposed framework and training strategy, our model is able to synthesize a photo-realistic output even for sketches of unseen classes.
We compare the proposed AODA to existing unpaired sketch-to-image generation approaches. Both qualitative and quantitative results show that our proposed method achieves significantly superior performance on both seen and unseen data.

The main contributions of this paper are three-fold: (1) We propose the adversarial open-domain adaptation (AODA) framework as the first attempt to solve the open-domain multi-class sketch-to-photo synthesis problem by learning to generate the missing freehand sketches. (2) We introduce an open-domain training strategy by considering certain fake sketches as real ones to reduce the generator's bias of synthesized sketches and leverage the generalization of adversarial domain adaptation, thus achieving more faithful generation for open-domain classes. (3) Our network provides, as a byproduct, a high-quality freehand sketch extractor for arbitrary photos. Extensive experiments and user studies on diverse datasets demonstrate that our model can faithfully synthesize realistic photos for different categories of open-domain freehand sketches. The source code and pre-trained models are available at \href{https://github.com/Mukosame/AODA}{https://github.com/Mukosame/AODA}.

\section{Related Work}
\noindent \textbf{Sketch-Based Image Synthesis} The goal of sketch-based image synthesis is to output a target image from a given sketch. Early works ~\cite{chen2009sketch2photo,eitz2011photosketcher,chen2012poseshop} regard freehand sketches as queries or constraints to retrieve each composition and stitch them into a picture. In recent years, an increasing number of works adopt GAN-based models~\cite{goodfellow2014generative} to learn pixel-wise translation between sketches and photos directly. \cite{zhu2017unpaired,li2019linestofacephoto,chen2020deepfacedrawing} train their networks with pairs of photos and corresponding edge maps due to the lack of real sketch data. However, the freehand sketches are usually distorted in shape compared with the target photo. Even when depicting the same object, the sketches from different users vary in appearance due to differences in their drawing skills and the levels of abstractness. To make the model applicable to freehand sketches, SketchyGAN~\cite{chen2018sketchygan} trained with both sketches and augmented edge maps. ContextualGAN~\cite{lu2018image} turns the image generation problem into an image completion problem: the network learns the joint distribution of sketch and image pairs and acquires the result by iteratively traversing the manifold. iSketchNFill~\cite{ghosh2019interactive} uses simple outlines to represent freehand sketches and generates photos from partial strokes with two-stage generators. Gao \etal~\cite{gao2020sketchycoco} applies two generators to synthesize the foreground and background respectively and proposes a novel GAN structure to encode the edge maps and corresponding photos into a shared latent space. The above works are supervised based on paired data. Liu \etal~\cite{liu2019unpaired} proposes a two-stage model for the unsupervised sketch-to-photo generation with reference images in a single class. Compared with these works, our problem setting is more challenging: we aim to learn the multi-class generation without supervision using paired data from an incomplete and heavily unbalanced dataset.

\noindent \textbf{Conditional Image Generation} Image generation can be controlled by class-condition~\cite{ghosh2019interactive,gao2020sketchycoco}, reference images~\cite{lu2018image,liu2019few,liu2019unpaired}, or specific semantic features~\cite{johnson2018image,park2019semantic,zhu2020sean}, \etc. The pioneering work cGAN~\cite{mirza2014conditional} combines the input noise with the condition for generator and discriminator. To help the generator synthesize images based on the input label, AC-GAN~\cite{odena2017conditional} makes the discriminator also predict the class labels. SGAN~\cite{odena2016semi} unifies the idea of discriminator and classifier by including the fake images as a new class. In this paper, we adopt a photo classifier that is jointly trained with the generator and discriminator to supervise the sketch-to-photo generator's training.

\section{Adversarial Open Domain Adaptation}
\label{sec:method}
First, we discuss the challenge of the open-domain generation problem and the limitation of previous methods in Section~\ref{subsec:challenge}. Then we introduce our proposed solution, including our AODA framework and the proposed training strategy in Section~\ref{sec:method_strategy}. 
\subsection{Challenge}
\label{subsec:challenge}
Unlike previous sketch-to-photo synthesis works~\cite{chen2018sketchygan,ghosh2019interactive} that can directly learn the mapping between the input sketch and its corresponding photo, during training, the sketches of open-domain classes are missing. To enable the network to learn to synthesize photos from sketches of both in-domain and open-domain classes, there are two ways to solve this problem: (1) training with extracted edge maps and (2) enriching the open-domain classes with synthesized sketches from a pre-trained photo-to-sketch extractor. We show the results of these two methods and discuss their limitations.

\begin{figure}[tbp]
\captionsetup[subfigure]{labelformat=empty}
\begin{center}
  \begin{subfigure}[b]{0.23\linewidth}
  \includegraphics[width=\linewidth]{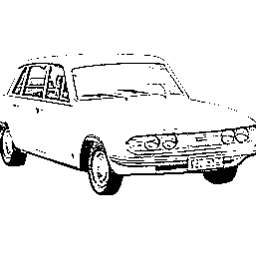}
  \end{subfigure}
  \begin{subfigure}[b]{0.23\linewidth}
  \includegraphics[width=\linewidth]{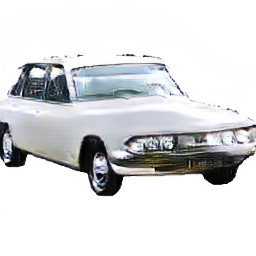}
  \end{subfigure}\hfill
\begin{subfigure}[b]{0.23\linewidth}
  \includegraphics[width=\linewidth]{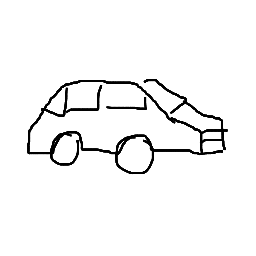}
  \end{subfigure}
  \begin{subfigure}[b]{0.23\linewidth}
  \includegraphics[width=\linewidth]{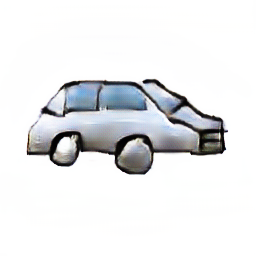}
  \end{subfigure}
  
  \begin{subfigure}[b]{0.23\linewidth}
  \includegraphics[width=\linewidth]{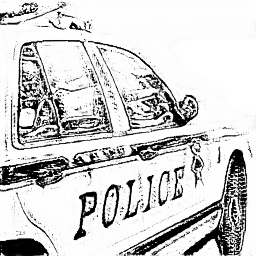}
  \subcaption{Edge}
  \end{subfigure}
  \begin{subfigure}[b]{0.23\linewidth}
  \includegraphics[width=\linewidth]{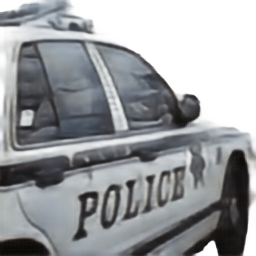}
  \subcaption{$\longrightarrow$ Output}
  \end{subfigure}\hfill
\begin{subfigure}[b]{0.23\linewidth}
  \includegraphics[width=\linewidth]{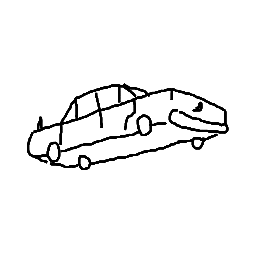}
  \subcaption{Real Sketch}
  \end{subfigure}
  \begin{subfigure}[b]{0.23\linewidth}
  \includegraphics[width=\linewidth]{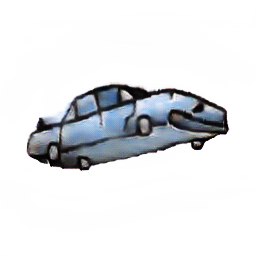}
  \subcaption{$\longrightarrow$ Output}
  \end{subfigure}
\end{center}
\caption{Results of photo synthesis from edge inputs and real sketch inputs generated by a model trained with xDoG edges and photos from the SketchyCOCO dataset~\cite{gao2020sketchycoco}. The left two columns show the xDoG inputs and their outputs, and the right two columns are the real freehand sketch inputs and the corresponding unsatisfactory outputs, which shows that the model simply trained with edges cannot rectify the distorted shapes of freehand sketches. }
 \label{fig:edge_failed_sketch}
\end{figure}

\noindent \textbf{Edge Maps.} Figure~\ref{fig:edge_failed_sketch} shows the results of a model trained on edges extracted by XDoG~\cite{winnemoller2012xdog}. While the model can generate vivid highlights and shadows and fine details from the edge inputs, the images generated from the actual freehand sketches are not that photo-realistic, but more like a colored drawing. This is because edges and freehand sketches are very different: freehand sketches are human abstractions of an object, usually with more deformations. The connections between the target photos and the input sketch are looser than with edges. Due to this domain gap, sketch-to-photo generators trained on the edge inputs usually cannot learn shape rectification, and thus fail to generalize to freehand sketches. 

\begin{figure}[tbp]
\captionsetup[subfigure]{labelformat=empty}
\begin{center}
  \begin{subfigure}[b]{0.23\linewidth}
  \includegraphics[width=\linewidth]{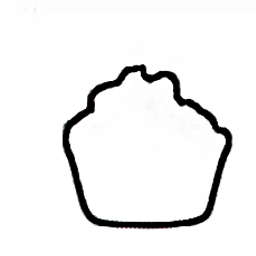}
  \end{subfigure}
  \begin{subfigure}[b]{0.23\linewidth}
  \includegraphics[width=\linewidth]{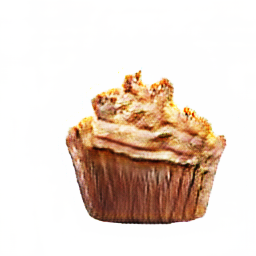}
  \end{subfigure}\hfill
   \begin{subfigure}[b]{0.23\linewidth}
  \includegraphics[width=\linewidth]{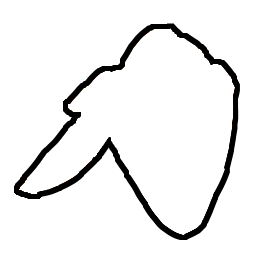}
  \end{subfigure}
  \begin{subfigure}[b]{0.23\linewidth}
  \includegraphics[width=\linewidth]{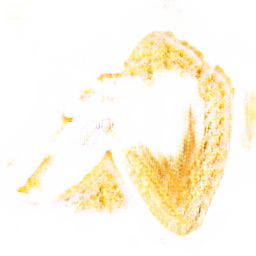}
  \end{subfigure}
  
  \begin{subfigure}[b]{0.23\linewidth}
  \includegraphics[width=\linewidth]{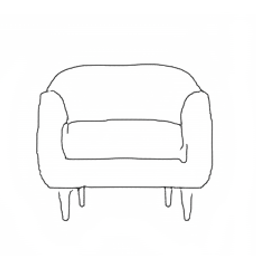}
  \subcaption{Fake Sketch}
  \end{subfigure}
  \begin{subfigure}[b]{0.23\linewidth}
  \includegraphics[width=\linewidth]{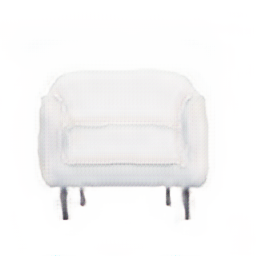}
  \subcaption{$\longrightarrow$ Output}
  \end{subfigure}\hfill
\begin{subfigure}[b]{0.23\linewidth}
  \includegraphics[width=\linewidth]{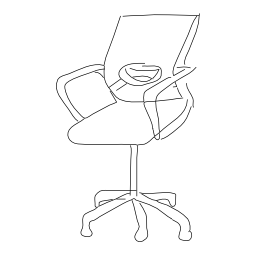}
  \subcaption{Real Sketch}
  \end{subfigure}
  \begin{subfigure}[b]{0.23\linewidth}
  \includegraphics[width=\linewidth]{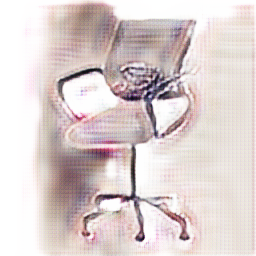}
  \subcaption{$\longrightarrow$ Output}
  \end{subfigure}
\end{center}
\caption{Results of photo synthesis from fake sketch inputs and real sketch inputs on Scribble~\cite{ghosh2019interactive} and QMUL-Sketch datasets~\cite{yu2016sketch,song2017deep,liu2019unpaired}. The outputs are generated by a model trained with synthesized sketches, and the setting remains the same as in~\cite{liu2019unpaired}, where the fake sketches are generated using a sketch extractor trained on the in-domain data. The left two columns show the fake sketch inputs and their outputs, and the right two columns are the real freehand sketch inputs and the corresponding unsatisfactory outputs. Comparing the outputs, we can see this training strategy makes the model fail to generalize on real sketches. }
 \label{fig:fixed_failed_sketch}
\end{figure}

\noindent \textbf{Synthesized sketches.} Another intuitive solution for open-domain generation is to enrich the training set of unseen classes $\mathcal{M}$ with sketches synthesized by a pre-trained photo-to-sketch generator~\cite{liu2019unpaired}. Figure~\ref{fig:fixed_failed_sketch} shows the result from a model trained with pre-extracted sketches on Scribble~\cite{ghosh2019interactive} and QMUL-Sketch dataset~\cite{yu2016sketch,song2017deep,liu2019unpaired}, where the photo-to-sketch extractor is trained with the in-domain classes of the training set. From the left two columns in Figure~\ref{fig:fixed_failed_sketch}, we can see that the model is able to generate photo-realistic outputs from synthesized sketches. However, it fails on real freehand sketches, as shown in the right two columns: even though it can generate the correct color and texture conditioned by the input label, it cannot understand the basic structure of real sketches (\eg distinguish the object from the background). The reason is, even though they are visually similar, the real and fake sketches are still distinguishable for the model. This strategy cannot guarantee that the model can generalize from the synthesized data to the real freehand sketches, especially for the multi-class generation. Thus, simply using the synthesized sketch to replace the missing freehand sketches cannot ensure photo-realistic generation. 

\subsection{Our Method}
\label{sec:method_strategy}
To solve this problem, we propose to learn the photo-to-sketch and sketch-to-photo translation jointly and to narrow the domain gap between the synthesized and real sketches.

\begin{figure*}[tbp]
    \centering
    \includegraphics[width=1.0 \textwidth]{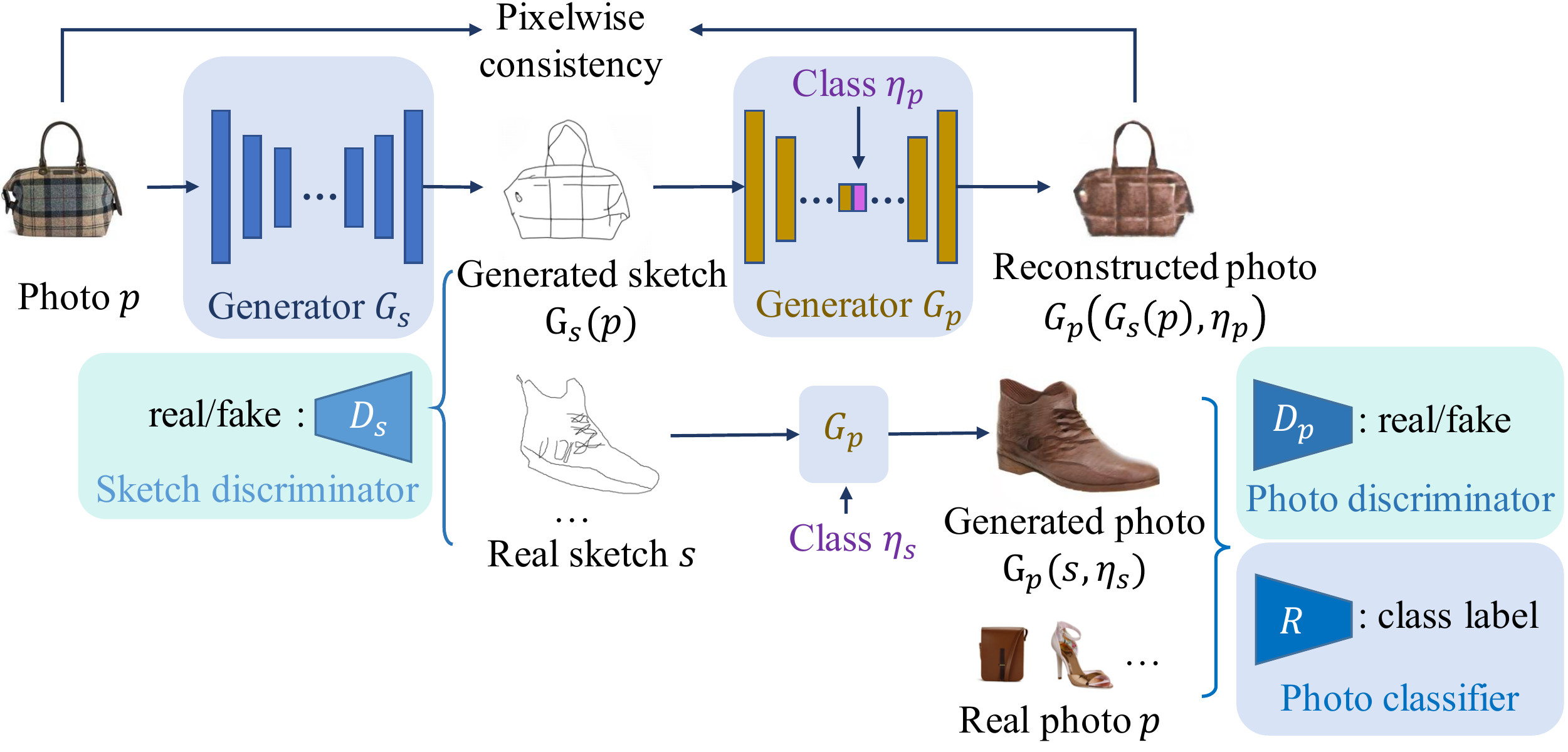}
    \caption{AODA framework overview. It has two generators $G_{s}: \text{photo} \rightarrow \text{sketch} $ and $G_{p}: \text{sketch} \rightarrow \text{photo}$ conditioned on the input label, and two discriminators $D_s$ and $D_p$ for the sketch and photo domains, respectively. In addition, we use a photo classifier $R$ to encourage $G_{p}$ to generate indistinguishable photos from the real ones of the same class.}
    \label{fig:structure}
\end{figure*}

\noindent \textbf{Framework.} As shown in Figure~\ref{fig:structure}, our framework mainly consists of the following parts: \textit{two generators}: a photo-to-sketch generator $G_{s}$, and a multi-class sketch-to-photo generator $G_{p}$ that takes sketch $s$ and class label $\eta_s$ as input; \textit{two discriminators} $D_s$ and $D_p$ that encourage the generators to synthesize outputs indistinguishable from the sketch domain $S$ and photo domain $P$, respectively; and a \textit{classifier $R$} that predicts class labels for both real photos $p$ and fake photos $G_{p}(s,\eta_s)$ to ensure that the output is truly conditioned on the input label $\eta_s$. Our AODA framework is trained with the unpaired sketch and photo data.

During the training process, $G_{p}$ extracts the sketch $G_{s}(p)$ from the given photo $p$. Then, the synthesized sketch $G_{s}(p)$ and the real sketch $s$ are sent to $G_{p}$ along with their labels $\eta_p$ and $\eta_s$, and turned into the reconstructed photo $G_{p}(G_{s}(p), \eta_p)$ and the synthesized photo $G_{p}(s, \eta_s)$, respectively. Note that we only send the sketch with its true label to ensure that $G_{p}$ learns the correct shape rectification from sketch to image domain for each class. The reconstructed photo is supposed to look similar to the original photo, which is imposed by a pixel-wise consistency loss. We do not add such a consistency constraint onto the sketch domain since we wish the synthesized sketches to be diverse. The generated photo is finally sent to the discriminator $D_p$ to ensure that it is photo-realistic, and to the classifier $R$ to ensure that it has the same perceptual features as the target class. In summary, the generator loss includes four parts: the adversarial loss of photo-to-sketch generation $\mathcal{L}_{G_{s}}$, the adversarial loss of sketch-to-photo translation $\mathcal{L}_{G_{p}}$, the pixel-wise consistency of photo reconstruction $\mathcal{L}_{pix}$, and the classification loss for the synthesized photo $\mathcal{L}_{\eta}$:
\begin{multline}
\label{eq:l_gan}
    \mathcal{L}_{GAN} = \lambda_s\mathcal{L}_{G_{s}}(G_{s}, D_s, p) + \lambda_p\mathcal{L}_{G_{p}}(G_{p},D_p,s,\eta_s) \\ + \lambda_{pix}\mathcal{L}_{pix}(G_{s},G_{p},p,\eta_p) + \lambda_{\eta}\mathcal{L}_{\eta}(R,G_{p},s,\eta_s).
\end{multline}

Please see our supplementary materials for more details. All parts of our framework are trained jointly from scratch. 

However, if we directly train the multi-class generator with the loss defined in Equation~\ref{eq:l_gan}, the training objective for open-domain classes $\mathcal{M}$ becomes the following form due to the missing sketches $s$:  
\begin{equation}
\label{eq:lgan_regress}
    \mathcal{L}_{GAN}^{\mathcal{M}} = \lambda_s\mathcal{L}_{G_{s}}(G_{s}, D_s, p) + \lambda_{pix}\mathcal{L}_{pix}(G_{s}, G_{p}, p, \eta_p),
\end{equation}
where $\eta_p \in \mathcal{M}$. As a result, the sketch-to-photo generator $G_{p}$ is solely supervised by the pixel-wise consistency. Since the commonly used $\mathcal{L}_1$ and $\mathcal{L}_2$ loss lead to the median and mean of pixels, respectively, this bias will make $G_{p}$ generate blurry photos for the open-domain classes.  

To solve this problem, we propose the random-mixed sampling strategy to minimize the domain gap between real and fake sketch inputs for the generator and to improve its output quality with the open-domain classes. 

\noindent \textbf{Random-mixed strategy.} This strategy aims to ``fool" the generator into treating fake sketches as real ones. Algorithm~\ref{alg:2_random} describes the detailed steps for the random-mixed sampling and modified optimization: $Pool$ denotes the buffer that stores the minibatch of sketch-label pairs. Querying the pool returns either the current minibatch or a previously stored one (and inserts the current minibatch in the pool) with a certain likelihood. $U$ denotes uniform sampling in the given range, and $t$ denotes the threshold that is set according to the ratio of open-domain classes and in-domain classes to match the photo data distribution.

\begin{algorithm}[ht]
\SetAlgoLined
\textbf{Input:} In training set $\mathcal{D}$, each minibatch contains photo set $p$, freehand sketch set $s$, the class label of photo $\eta_{p}$, and the class label of sketch $\eta_{s}$\;
 \For{number of training iterations}{
  $s_{fake}\leftarrow G_{s}(p)$\; 
  $s_{c} \leftarrow s$\;
  $\eta_{c} \leftarrow \eta_{s}$\;
  \If {$t<u\sim U(0, 1)$}{
   $s_{c},\eta_{c} \leftarrow pool$.query($s_{fake}$, $\eta_{p}$)\; 
   }
   $p_{rec} \leftarrow G_{p}(s_{fake},\eta_{p})$\;
   $p_{fake}\leftarrow G_{p}(s, \eta_{s})$

  Calculate $\mathcal{L}_{GAN}$ with $(p, s_c, p_{rec}, \eta_c)$ and update $G_{s}$ and $G_{p}$\;
  Calculate $\mathcal{L}_{D_s}(s,s_{fake})$ and $\mathcal{L}_{D_s}(p,p_{fake})$, update $D_s$ and $D_p$\;
  Calculate $\mathcal{L}_R(p,p_{fake},\eta_p,\eta_s)$ and update the classifier.
 }
 \caption{Minibatch Random-Mixed Sampling and Updating}
 \label{alg:2_random}
\end{algorithm}

One key operation of this strategy is to construct pseudo sketches for $G_{p}$ by randomly mixing the synthesized sketches with real ones in a batch-wise manner. In this step, the pseudo sketches are treated as the real ones by the generator. Thus, the open-domain classes' $\mathcal{L}_{GAN}^{\mathcal{M}}$ becomes:
\begin{multline}
    \mathcal{L}_{GAN}^{\mathcal{M}} = \lambda_s\mathcal{L}_{G_{s}}(G_{s}, D_s, p) + \lambda_p\mathcal{L}_{G_{p}}(G_{p},D_p,s_{fake},\eta_p) \\ + \lambda_{pix}\mathcal{L}_{pix}(G_{s},G_{p},p,\eta_p) + \lambda_{\eta}\mathcal{L}_{\eta}(R,G_{p},s_{fake},\eta_p),
\end{multline}
where $\eta_p \in \mathcal{M}$. Another key of the strategy is on optimization: the sampling strategy is only for $G_{p}$. The classifier and discriminators are still updated with real/fake data to guarantee their discriminative powers. 

The random mixing operation is blind to in-domain and open-domain classes. As a result, the training sketches include both real and pseudo sketches from all categories. By including pseudo sketches from both the in-domain and open-domain classes, it would further enforce the sketch-to-image generator to ignore the domain gap in the inputs and synthesize realistic photos from both real and fake sketches. Note that since $ G_{s}$'s parameters are consistently updated during training, the pseudo sketches also change for each batch. Moreover, the pseudo sketch-label pairs are acquired from a history of generated sketches and their labels rather than the latest produced ones by $G_{s}$. We maintain a buffer that stores the 50 previously added minibatch of sketch-label pairs \cite{shrivastava2017learning,zhu2017unpaired}. 

Mixing real sketches with fake ones can be regarded as an online data augmentation technique for training $G_{p}$. Compared with augmentation using edges, $G_{s}$ can learn the distortions from real freehand sketches by approaching the real data distribution~\cite{goodfellow2014generative,ledig2017photo,zhou2019kernel}, and enable $G_{p}$ to learn shape rectification on the fly. Benefiting from the joint training mechanism, as the training progresses, the sketches generated by $G_{s}$ change epoch by epoch. The loose consistency constraint on sketch generation further increases the diverseness of the sketch data in the open-domain. Compared with using pre-extracted sketches, the open-domain buffer maintains a broad spectrum of sketches: from the very coarse ones generated in early epochs to the more human-like sketches in later epochs as $G_{s}$ converges. 


\section{Experiments}

\subsection{Experiment Setup}
\label{sec:exp_setup}
\noindent \textbf{Dataset.} We train and evaluate the performance of sketch-to-photo synthesis methods on two datasets: Scribble~\cite{ghosh2019interactive} (10 classes), and SketchyCOCO~\cite{gao2020sketchycoco} (14 classes of objects). 

Scribble contains ten object classes with photos and simple outline sketches. Six out of ten classes have similar round outlines, which imposes more stringent requirements on the network: whether it can generate the correct structure and texture conditioned on the input class label. In the open-domain setting, we only have the sketches of four classes for training: \textit{pineapple}, \textit{cookie}, \textit{orange}, and \textit{watermelon}, which means that $60\%$ of the classes are open-domain. 


SketchyCOCO includes 14 object classes, where the sketches are collected from the Sketchy dataset~\cite{sangkloy2016sketchy}, TU-Berlin dataset~\cite{eitz2012humans}, and \textit{Quick! Draw} dataset~\cite{ha2017neural}. The 14,081 photos for each object class are segmented from the natural images of COCO Stuff~\cite{caesar2018coco} under unconstrained conditions, thereby making it more difficult for existing methods to map the freehand sketches to the photo domain. The two open-domain classes are: \textit{sheep} and \textit{giraffe}. 

\noindent \textbf{Metrics.} We quantitatively evaluate the generation results with three different metrics: 1) Fr\'{e}chet Inception Distance (FID)~\cite{heusel2017gans} that measures the feature similarity between generated and real image sets. A Low FID score means the generated images are less different from the real ones and thus have high fidelity; 2) Classification Accuracy (Acc)~\cite{ashual2019specifying} of generated images with a pre-trained classifier in the same manner as~\cite{ghosh2019interactive,gao2020sketchycoco}. Higher accuracy indicates better image realism; 3) User Preference Study (Human): we show the participants a given sketch and the class label, and ask them to pick one photo with the best quality and realism from generated results. We randomly sample 31 groups of images. For each evaluation, we shuffle the options and show them to 25 users. We collect 775 answers in total.

\subsection{Sketch-to-Photo Synthesis}
\subsubsection{Comparison to Other Methods}
\label{sec:exp_comp}
To better illustrate the effectiveness of our proposed solution, here we adopt CycleGAN~\cite{zhu2017unpaired} as the baseline in building our network and include the original CycleGAN in the following comparison. To make it be able to accept sketch class labels, we modified the sketch-to-photo translator to be a conditional generator. Besides, we also compare to a recent work EdgeGAN~\cite{gao2020sketchycoco} on each dataset. We mark the open-domain sketch with a \textcolor{blue}{$\bigstar$} for better visualization.

\noindent \textbf{Scribble.}  Figure~\ref{fig:scribble_results} shows the qualitative results of (a) CycleGAN, (b) conditional CycleGAN, (c) conditional CycleGAN with classification loss, (d) EdgeGAN and our method, where the bottom three rows are open-domain. The original CycleGAN exhibits mode collapse and synthesizes identical textures for all categories, probably due to the fact that the sketches in the Scribble dataset barely imply their class labels. This problem is alleviated in (b). Still, it fails to synthesize natural photos for some categories due to the gap between open-domain and in-domain data. Such a domain gap is even worse in (c), where the in-domain result is with realistic but wrong texture, and the open-domain results are texture-less. This might be because that classifier implicitly increases the domain gap while maximizing the class discrepancy. Thus, we do not include this model for comparison on the SketchyCOCO dataset. Compared with (d), our results are more consistent with the input sketch shape, demonstrating that our model is better at understanding the composition in sketches and learning more faithful shape rectification in sketch-to-photo domain mapping.

\newcommand{\scribblewidth}{0.15} 
\begin{figure}[tbp]
\captionsetup[subfigure]{labelformat=empty}
\begin{center}
  \begin{subfigure}[b]{\scribblewidth\linewidth}
  \includegraphics[width=\linewidth]{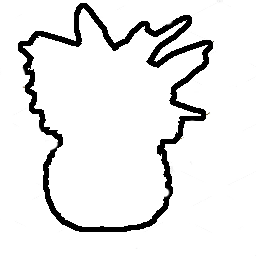}
  \end{subfigure}
  \begin{subfigure}[b]{\scribblewidth\linewidth}
  \includegraphics[width=\linewidth]{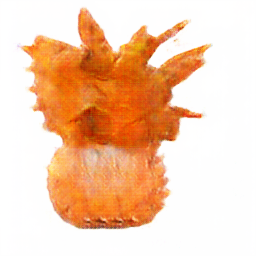}
  \end{subfigure}
   \begin{subfigure}[b]{\scribblewidth\linewidth}
  \includegraphics[width=\linewidth]{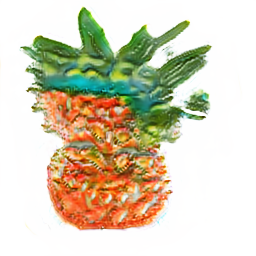}
  \end{subfigure}
   \begin{subfigure}[b]{\scribblewidth\linewidth}
  \includegraphics[width=\linewidth]{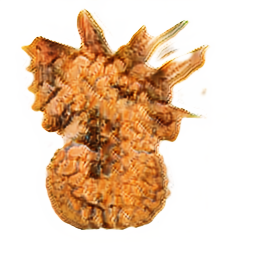}
  \end{subfigure}
 \begin{subfigure}[b]{\scribblewidth\linewidth}
  \includegraphics[width=\linewidth]{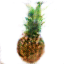}
  \end{subfigure}
  \begin{subfigure}[b]{\scribblewidth\linewidth}
  \includegraphics[width=\linewidth]{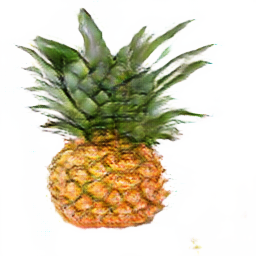}
  \end{subfigure}
  
  \begin{subfigure}[b]{\scribblewidth\linewidth}
  \includegraphics[width=\linewidth]{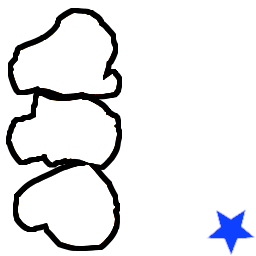}
  \end{subfigure}
  \begin{subfigure}[b]{\scribblewidth\linewidth}
  \includegraphics[width=\linewidth]{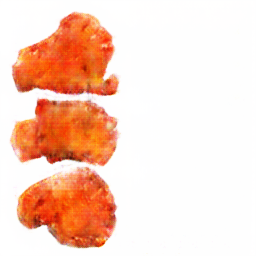}
  \end{subfigure}
   \begin{subfigure}[b]{\scribblewidth\linewidth}
  \includegraphics[width=\linewidth]{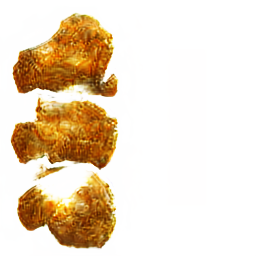}
  \end{subfigure}
   \begin{subfigure}[b]{\scribblewidth\linewidth}
  \includegraphics[width=\linewidth]{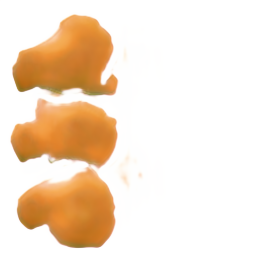}
  \end{subfigure}
 \begin{subfigure}[b]{\scribblewidth\linewidth}
  \includegraphics[width=\linewidth]{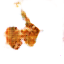}
  \end{subfigure}
  \begin{subfigure}[b]{\scribblewidth\linewidth}
  \includegraphics[width=\linewidth]{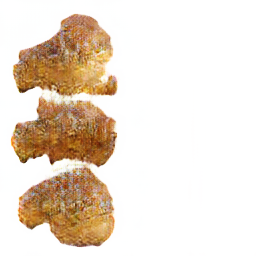}
  \end{subfigure}
  
  \begin{subfigure}[b]{\scribblewidth\linewidth}
  \includegraphics[width=\linewidth]{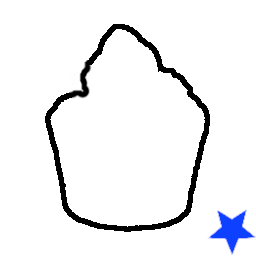}
  \end{subfigure}
  \begin{subfigure}[b]{\scribblewidth\linewidth}
  \includegraphics[width=\linewidth]{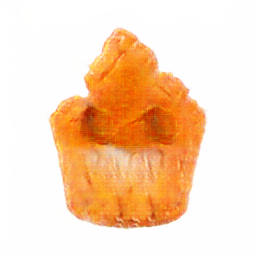}
  \end{subfigure}
   \begin{subfigure}[b]{\scribblewidth\linewidth}
  \includegraphics[width=\linewidth]{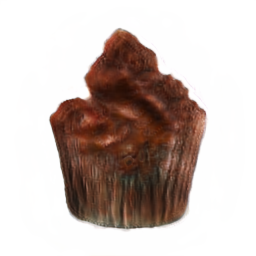}
  \end{subfigure}
   \begin{subfigure}[b]{\scribblewidth\linewidth}
  \includegraphics[width=\linewidth]{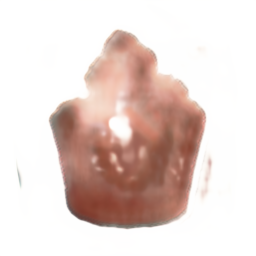}
  \end{subfigure}
 \begin{subfigure}[b]{\scribblewidth\linewidth}
  \includegraphics[width=\linewidth]{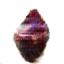}
  \end{subfigure}
  \begin{subfigure}[b]{\scribblewidth\linewidth}
  \includegraphics[width=\linewidth]{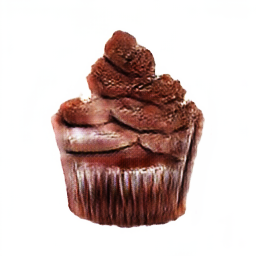}
  \end{subfigure}
  
  \begin{subfigure}[b]{\scribblewidth\linewidth}
  \includegraphics[width=\linewidth]{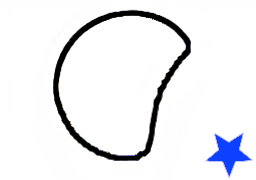}
  \subcaption{Input}
  \end{subfigure}
  \begin{subfigure}[b]{\scribblewidth\linewidth}
  \includegraphics[width=\linewidth]{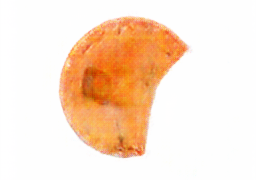}
  \subcaption{(a)}
  \end{subfigure}
\begin{subfigure}[b]{\scribblewidth\linewidth}
  \includegraphics[width=\linewidth]{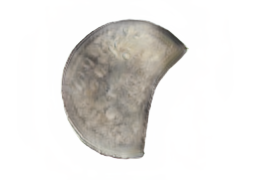}
  \subcaption{(b)}
  \end{subfigure}
\begin{subfigure}[b]{\scribblewidth\linewidth}
  \includegraphics[width=\linewidth]{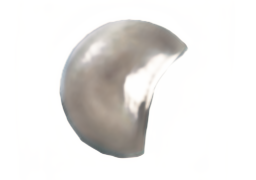}
  \subcaption{(c)}
  \end{subfigure}
\begin{subfigure}[b]{\scribblewidth\linewidth}
  \includegraphics[width=\linewidth]{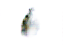}
  \subcaption{(d)}
  \end{subfigure}
  \begin{subfigure}[b]{\scribblewidth\linewidth}
  \includegraphics[width=\linewidth]{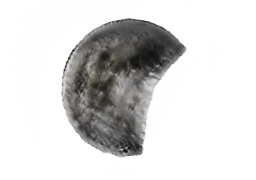}
  \subcaption{Ours}
  \end{subfigure}
\end{center}
\caption{Results on Scribble dataset~\cite{ghosh2019interactive}. We mark the open-domain inputs with \textcolor{blue}{$\bigstar$}. The following columns are outputs of (a) CycleGAN~\cite{zhu2017unpaired}, (b) conditional CycleGAN, (c) classifier+(b), (d) EdgeGAN~\cite{gao2020sketchycoco}, and ours.}
 \label{fig:scribble_results}
\end{figure}

\noindent \textbf{SketchyCOCO} The qualitative results are shown in Figure~\ref{fig:sketchycoco_results}, where the top two rows are of in-domain categories, and the bottom two are open-domain. The photos generated by CycleGAN suffer from mode collapse. As shown in column (b), conditional CycleGAN cannot generate vivid textures for open-domain categories. Compared with EdgeGAN in (c), the poses in our generated photos are more faithful to the input sketches. 

\newcommand{\cocowidth}{0.18} 
\begin{figure}[tbp]
\captionsetup[subfigure]{labelformat=empty}
\begin{center}
  \begin{subfigure}[b]{\cocowidth\linewidth}
  \includegraphics[width=\linewidth]{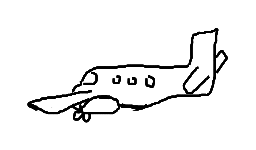}
  \end{subfigure}
  \begin{subfigure}[b]{\cocowidth\linewidth}
  \includegraphics[width=\linewidth]{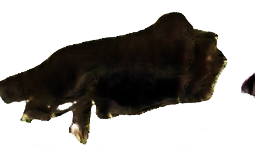}
  \end{subfigure}
  \begin{subfigure}[b]{\cocowidth\linewidth}
  \includegraphics[width=\linewidth]{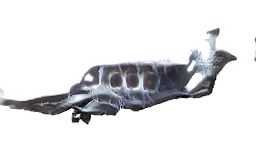}
  \end{subfigure}
   \begin{subfigure}[b]{\cocowidth\linewidth}
  \includegraphics[width=\linewidth]{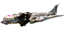}
  \end{subfigure}
  \begin{subfigure}[b]{\cocowidth\linewidth}
  \includegraphics[width=\linewidth]{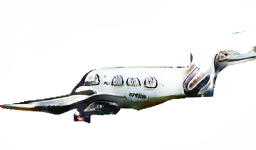}
  \end{subfigure}
  
\begin{subfigure}[b]{\cocowidth\linewidth}
  \includegraphics[width=\linewidth]{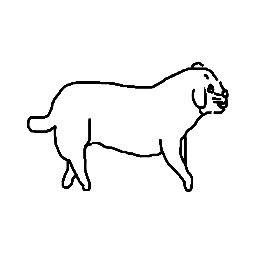}
  \end{subfigure}
  \begin{subfigure}[b]{\cocowidth\linewidth}
  \includegraphics[width=\linewidth]{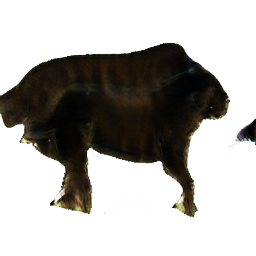}
  \end{subfigure}
  \begin{subfigure}[b]{\cocowidth\linewidth}
  \includegraphics[width=\linewidth]{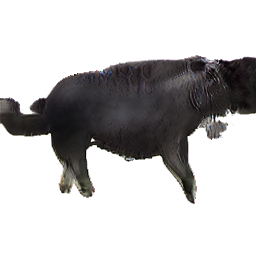}
  \end{subfigure}
   \begin{subfigure}[b]{\cocowidth\linewidth}
  \includegraphics[width=\linewidth]{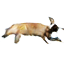}
  \end{subfigure}
  \begin{subfigure}[b]{\cocowidth\linewidth}
  \includegraphics[width=\linewidth]{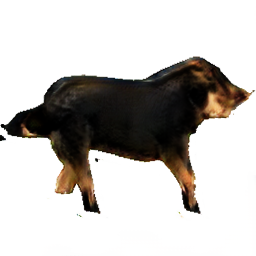}
  \end{subfigure}
  \begin{subfigure}[b]{\cocowidth\linewidth}
  \includegraphics[width=\linewidth]{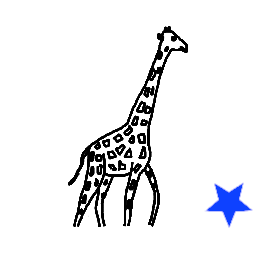}
  \end{subfigure}
  \begin{subfigure}[b]{\cocowidth\linewidth}
  \includegraphics[width=\linewidth]{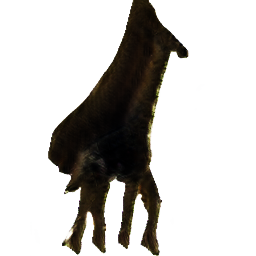}
  \end{subfigure}
  \begin{subfigure}[b]{\cocowidth\linewidth}
  \includegraphics[width=\linewidth]{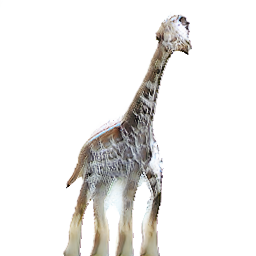}
  \end{subfigure}
   \begin{subfigure}[b]{\cocowidth\linewidth}
  \includegraphics[width=\linewidth]{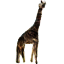}
  \end{subfigure}
  \begin{subfigure}[b]{\cocowidth\linewidth}
  \includegraphics[width=\linewidth]{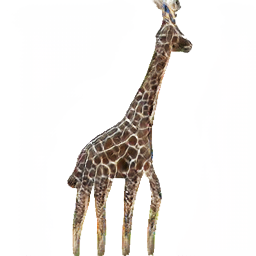}
  \end{subfigure}
  
  \begin{subfigure}[b]{\cocowidth\linewidth}
  \includegraphics[width=\linewidth]{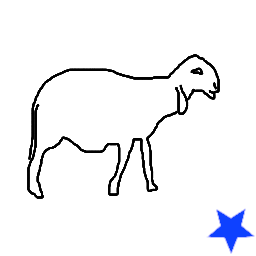}
  \subcaption{Input}
  \end{subfigure}
  \begin{subfigure}[b]{\cocowidth\linewidth}
  \includegraphics[width=\linewidth]{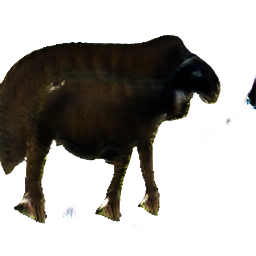}
  \subcaption{(a)}
  \end{subfigure}
 \begin{subfigure}[b]{\cocowidth\linewidth}
  \includegraphics[width=\linewidth]{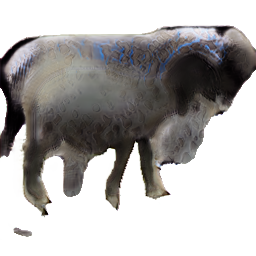}
  \subcaption{(b)}
  \end{subfigure}
\begin{subfigure}[b]{\cocowidth\linewidth}
  \includegraphics[width=\linewidth]{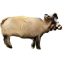}
  \subcaption{(c)}
  \end{subfigure}
  \begin{subfigure}[b]{\cocowidth\linewidth}
  \includegraphics[width=\linewidth]{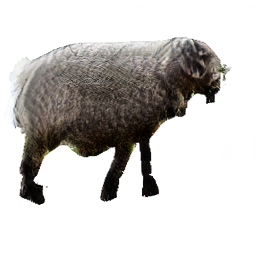}
  \subcaption{Ours}
  \end{subfigure}
\end{center}
\caption{Results on SketchyCOCO dataset~\cite{gao2020sketchycoco} for the compared methods: (a) CycleGAN~\cite{zhu2017unpaired}, (b) conditional CycleGAN, (c) EdgeGAN~\cite{gao2020sketchycoco}, and ours. The open-domain inputs are marked with \textcolor{blue}{$\bigstar$}. }
 \label{fig:sketchycoco_results}
\end{figure}

\begin{table*}[t]

\centering
\resizebox{\textwidth}{!}{
\begin{tabular}{c|c|ccc|ccc|ccc|ccc}
\hline
Dataset     & Method      & \multicolumn{3}{c|}{CycleGAN~\cite{zhu2017unpaired}}     & \multicolumn{3}{c|}{conditional CycleGAN}     & \multicolumn{3}{c|}{EdgeGAN~\cite{gao2020sketchycoco}}      &    \multicolumn{3}{c}{Ours}                  \\
            & Metric      & \multicolumn{1}{c}{full} & \multicolumn{1}{c}{in-domain} & \multicolumn{1}{c|}{open-domain} & \multicolumn{1}{c}{full} & \multicolumn{1}{c}{in-domain} & \multicolumn{1}{c|}{open-domain} & \multicolumn{1}{c}{full} & \multicolumn{1}{c}{in-domain} & \multicolumn{1}{c|}{open-domain} & full                      & in-domain                      & open-domain                     \\ \hline
\hline
Scribble    & FID $\downarrow$      & 279.5                    & 252.7                    & 355.9                      & 213.6                    & 210.9                    & 253.6                      & 259.7                    & 256.3                    & 298.5                      & \textbf{209.5} & \textbf{204.6} & \textbf{252.8}  \\
            & Acc (\%) $\uparrow$ & 16.0                       & 30.0                       & 6.7                        & 68.0                       & 70.0                       & 66.7                       & \textbf{100.0}                      & \textbf{100.0}                      & \textbf{100.0}                        & \textbf{100.0}   & \textbf{100.0}   & \textbf{100.0}    \\
            & Human (\%) $\uparrow$&    5.60  & 1.00    &   8.67    &  19.20   &  17.00  & 20.67    &  25.20   &  17.00  &  30.67     &     \textbf{48.80}                      &         \textbf{65.00}                  &         \textbf{38.00}                   \\
\hline
SketchyCOCO & FID $\downarrow$       & 201.7                    & 218.7                    & 237.2                      & 124.3                    & 138.7                    & 171.6                      & 169.7                    & 177.8                    & 221                        & \textbf{114.8} & \textbf{128.4} & \textbf{139.2}  \\
            & Acc (\%) $\uparrow$ & 8.4                      & 10.8                     & 1.9                        & 57.0                       & 58.7                     & 52.4                       & 75.8                     & 68.8                     & 98.3                       & \textbf{78.3}  & \textbf{70.5}  & \textbf{100.0}    \\
            & Human (\%) $\uparrow$&  0.36    & 0.00   &   0.67    &  5.09   & 5.60   &   4.67  &  22.55   & 32.00   &   14.67    &   \textbf{72.00}                        &       \textbf{59.20}                    &         \textbf{82.67}                   \\ 
\hline
\end{tabular}
}
\caption{Quantitative evaluation and user study on Scribble and SketchyCOCO datasets. We show the full testset results, in-domain results, and open-domain results, respectively. Best results are shown in \textbf{bold}.}

\label{tab:results}
\end{table*}

The quantitative results for the three datasets are summarized in Table~\ref{tab:results}. We can see that our model is preferred by more users than the other compared methods, and achieves the best results in terms of the FID score and classification accuracy on all datasets. These results confirm our observations of the qualitative outputs, as discussed above. Besides, we have an interesting observation: compared with the baseline CycleGAN and conditional CycleGAN, our random-mixed strategy improves not only the open-domain results, but also in-domain results. We find a possible explanation from \cite{tao2020alleviation}: the ``fake-as-real" operation can effectively alleviate the gradient exploding issue during GAN training and result in a more faithful generated distribution.


\subsubsection{Robustness}
\label{sec:exp_robust}
We test our sketch-to-photo generator's robustness to the inputs and show the visual results in Figure~\ref{fig:robust}: the left two columns show partial sketches that are generated by removing some strokes from the original one, and the right two columns are enriched sketches that are generated by adding extra strokes to the original ones. The original sketch from the SketchyCOCO~\cite{gao2020sketchycoco} test set and its output are shown in the middle column. Our model can synthesize realistic airplanes, even when the image composition is changed by adding or removing strokes. 

\newcommand{\robustwidth}{0.18} 
\begin{figure}[tbp]
\captionsetup[subfigure]{labelformat=empty}
\begin{center}
\begin{subfigure}[b]{\robustwidth\linewidth}
  \includegraphics[width=\linewidth]{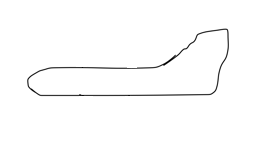}
  \end{subfigure}
  \begin{subfigure}[b]{\robustwidth\linewidth}
  \includegraphics[width=\linewidth]{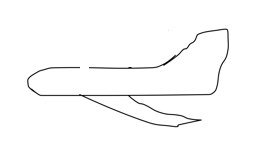}
  \end{subfigure}  
  \begin{subfigure}[b]{\robustwidth\linewidth}
  \includegraphics[width=\linewidth]{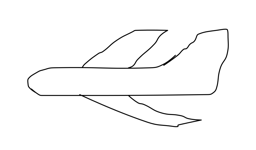}
  \end{subfigure}
  \begin{subfigure}[b]{\robustwidth\linewidth}
  \includegraphics[width=\linewidth]{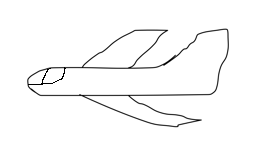}
  \end{subfigure}
\begin{subfigure}[b]{\robustwidth\linewidth}
  \includegraphics[width=\linewidth]{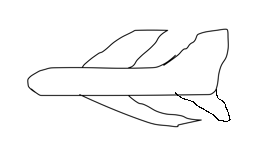}
  \end{subfigure}

\begin{subfigure}[b]{\robustwidth\linewidth}
  \includegraphics[width=\linewidth]{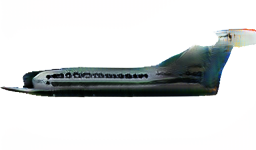}
  \subcaption{$--$}
  \end{subfigure}
 \begin{subfigure}[b]{\robustwidth\linewidth}
  \includegraphics[width=\linewidth]{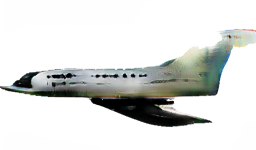}
  \subcaption{$-$}
  \end{subfigure}  
\begin{subfigure}[b]{\robustwidth\linewidth}
  \includegraphics[width=\linewidth]{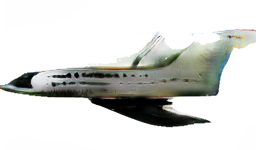}
  \subcaption{Original}
  \end{subfigure}
  \begin{subfigure}[b]{\robustwidth\linewidth}
  \includegraphics[width=\linewidth]{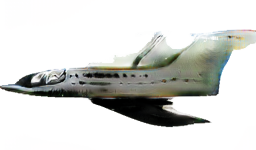}
  \subcaption{$+$}
  \end{subfigure}
\begin{subfigure}[b]{\robustwidth\linewidth}
  \includegraphics[width=\linewidth]{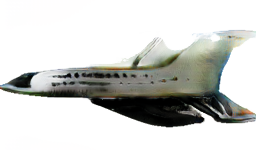}
  \subcaption{$++$}
  \end{subfigure}
\end{center}
\caption{Our model works well for the sketches that are modified by removing strokes (left two columns) and adding strokes (right two columns).}
 \label{fig:robust}
\end{figure}

\subsection{Photo-to-Sketch Synthesis}
\label{sec:p2s_syn}
As a byproduct, our network can also provide a high-quality freehand sketch generator $G_{s}$ for a given photo \cite{pang2018deep,yi2019apdrawinggan,kampelmuhler2020synthesizing}. We run our sketch extractor on COCO objects (top two rows) and web images (bottom two rows) and display the results in Figure~\ref{fig:photo_to_sketch}. Our model can generate different styles of freehand sketches like human drawers beyond the edge map of a photo, even for unseen objects.

\newcommand{\ptswidth}{0.22} 
\begin{figure}[tbp]
\captionsetup[subfigure]{labelformat=empty}
\begin{center}
  \begin{subfigure}[b]{\ptswidth\linewidth}
  \includegraphics[width=\linewidth]{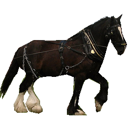}
  \end{subfigure}
  \begin{subfigure}[b]{\ptswidth\linewidth}
  \includegraphics[width=\linewidth]{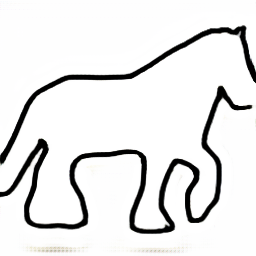}
  \end{subfigure}
\begin{subfigure}[b]{\ptswidth\linewidth}
  \includegraphics[width=\linewidth]{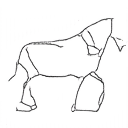}
  \end{subfigure}
  \begin{subfigure}[b]{\ptswidth\linewidth}
  \includegraphics[width=\linewidth]{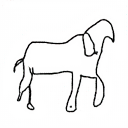}
  \end{subfigure}
  
  \begin{subfigure}[b]{\ptswidth\linewidth}
  \includegraphics[width=\linewidth]{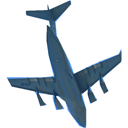}
  \end{subfigure}
  \begin{subfigure}[b]{\ptswidth\linewidth}
  \includegraphics[width=\linewidth]{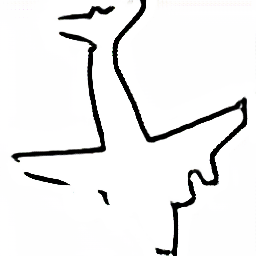}
  \end{subfigure}
\begin{subfigure}[b]{\ptswidth\linewidth}
  \includegraphics[width=\linewidth]{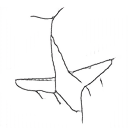}
  \end{subfigure}
  \begin{subfigure}[b]{\ptswidth\linewidth}
  \includegraphics[width=\linewidth]{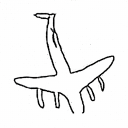}
  \end{subfigure}
  
  \begin{subfigure}[b]{\ptswidth\linewidth}
  \includegraphics[width=\linewidth]{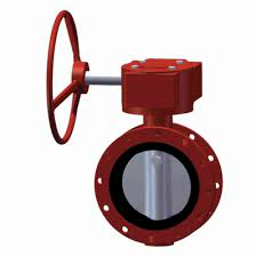}
  \end{subfigure}
  \begin{subfigure}[b]{\ptswidth\linewidth}
  \includegraphics[width=\linewidth]{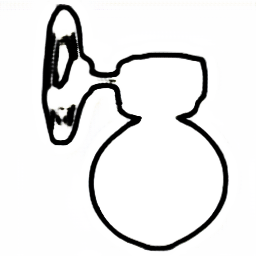}
  \end{subfigure}
\begin{subfigure}[b]{\ptswidth\linewidth}
  \includegraphics[width=\linewidth]{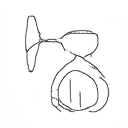}
  \end{subfigure}
  \begin{subfigure}[b]{\ptswidth\linewidth}
  \includegraphics[width=\linewidth]{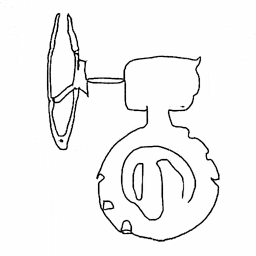}
  \end{subfigure}
  
  \begin{subfigure}[b]{\ptswidth\linewidth}
  \includegraphics[width=\linewidth]{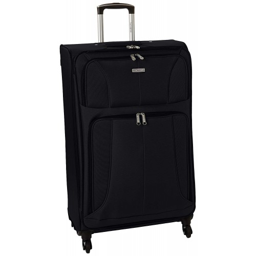}
  \subcaption{Input}
  \end{subfigure}
  \begin{subfigure}[b]{\ptswidth\linewidth}
  \includegraphics[width=\linewidth]{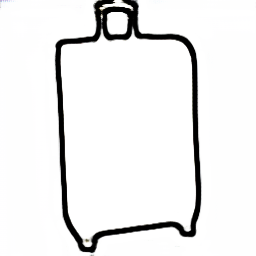}
  \subcaption{Scribble Style}
  \end{subfigure}
\begin{subfigure}[b]{\ptswidth\linewidth}
  \includegraphics[width=\linewidth]{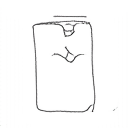}
  \subcaption{QMUL Style}
  \end{subfigure}
  \begin{subfigure}[b]{\ptswidth\linewidth}
  \includegraphics[width=\linewidth]{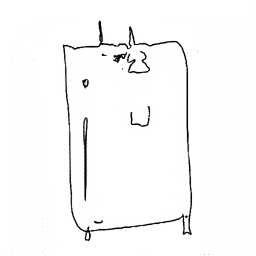}
  \subcaption{Sketchy Style}
  \end{subfigure}
\end{center}
\caption{Photo-based sketch synthesis results. Given a photo input, as shown in the first column, our photo-to-sketch generator can translate it into sketches in different styles. Our model is able to generate freehand sketches like human drawers on both seen classes and unseen classes.}
 \label{fig:photo_to_sketch}
\end{figure}

Characterized by the joint training, the weights of the photo-to-sketch generator are constantly updated as the training progresses. As a result, the sketches generated by $G_{s}$ change epoch by epoch. Figure~\ref{fig:p2s_epoch} shows the extracted sketches at different epochs. The changing sketches increase the diverseness of the sketch, and thus can further augment the data and help the sketch-to-photo generator to better generalize to various freehand sketch inputs.

\newcommand{\ptsewidth}{0.18} 
\begin{figure}[tbp]
\captionsetup[subfigure]{labelformat=empty}
\begin{center}
  \begin{subfigure}[b]{\ptsewidth\linewidth}
  \includegraphics[width=\linewidth]{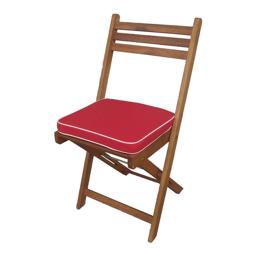}
  \end{subfigure}
  \begin{subfigure}[b]{\ptsewidth\linewidth}
  \includegraphics[width=\linewidth]{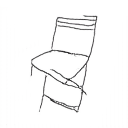}
  \end{subfigure}
\begin{subfigure}[b]{\ptsewidth\linewidth}
  \includegraphics[width=\linewidth]{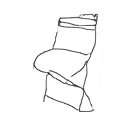}
  \end{subfigure}
  \begin{subfigure}[b]{\ptsewidth\linewidth}
  \includegraphics[width=\linewidth]{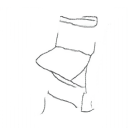}
  \end{subfigure}
  \begin{subfigure}[b]{\ptsewidth\linewidth}
  \includegraphics[width=\linewidth]{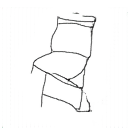}
  \end{subfigure}

\begin{subfigure}[b]{\ptsewidth\linewidth}
  \includegraphics[width=\linewidth]{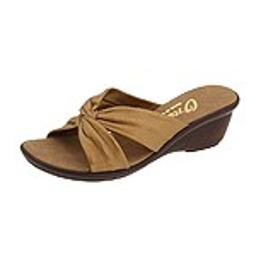}
  \end{subfigure}
  \begin{subfigure}[b]{\ptsewidth\linewidth}
  \includegraphics[width=\linewidth]{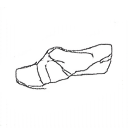}
  \end{subfigure}
\begin{subfigure}[b]{\ptsewidth\linewidth}
  \includegraphics[width=\linewidth]{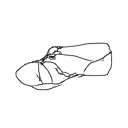}
  \end{subfigure}
  \begin{subfigure}[b]{\ptsewidth\linewidth}
  \includegraphics[width=\linewidth]{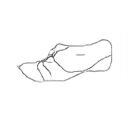}
  \end{subfigure}
  \begin{subfigure}[b]{\ptsewidth\linewidth}
  \includegraphics[width=\linewidth]{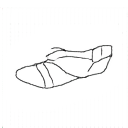}
  \end{subfigure}
  
  \begin{subfigure}[b]{\ptsewidth\linewidth}
  \includegraphics[width=\linewidth]{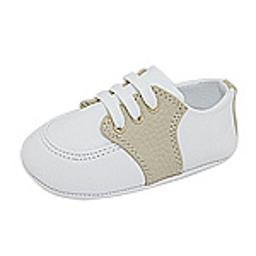}
  \subcaption{Input}
  \end{subfigure}
  \begin{subfigure}[b]{\ptsewidth\linewidth}
  \includegraphics[width=\linewidth]{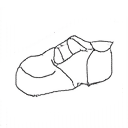}
  \subcaption{epoch=10}
  \end{subfigure}
\begin{subfigure}[b]{\ptsewidth\linewidth}
  \includegraphics[width=\linewidth]{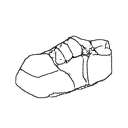}
  \subcaption{20}
  \end{subfigure}
  \begin{subfigure}[b]{\ptsewidth\linewidth}
  \includegraphics[width=\linewidth]{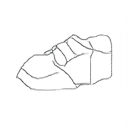}
  \subcaption{30}
  \end{subfigure}
  \begin{subfigure}[b]{\ptsewidth\linewidth}
  \includegraphics[width=\linewidth]{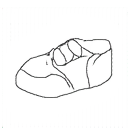}
  \subcaption{40}
  \end{subfigure}
\end{center}
\caption{Photo-based sketch synthesis results at different epochs. Given an input shown in the first column, synthesized sketches from our model change at different epochs.}
 \label{fig:p2s_epoch}
\end{figure}

\subsection{Ablation Study}
\label{sec:exp_abl}

\noindent \textbf{Effectiveness of AODA} To illustrate the effect of the proposed open-domain training strategy, we simplify the dataset to two classes, including the in-domain class \textit{pineapple} and the open-domain class \textit{strawberry}. We compare four models: (a) baseline CycleGAN without classifier or sampling strategy; (b) AODA framework without sampling strategy; (c) AODA trained with synthesized open-domain sketches and real in-domain sketches; (d) AODA trained with the random-mixed sampling strategy as described in Algorithm~\ref{alg:2_random}. The results are shown in Figure~\ref{fig:abl_strategy},

\newcommand{\strwidth}{0.19} 
\begin{figure}[tbp]
\captionsetup[subfigure]{labelformat=empty}
\begin{center}
  \begin{subfigure}[b]{\strwidth\linewidth}
  \includegraphics[width=\linewidth]{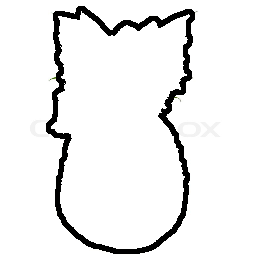}
  \end{subfigure}
  \begin{subfigure}[b]{\strwidth\linewidth}
  \includegraphics[width=\linewidth]{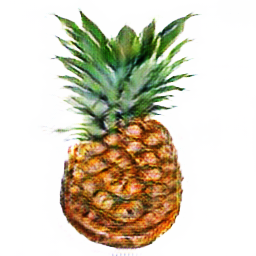}
  \end{subfigure}
  \begin{subfigure}[b]{\strwidth\linewidth}
  \includegraphics[width=\linewidth]{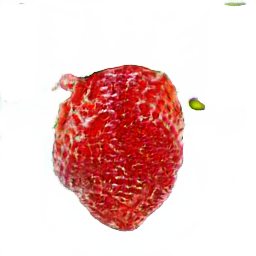}
  \end{subfigure}
   \begin{subfigure}[b]{\strwidth\linewidth}
  \includegraphics[width=\linewidth]{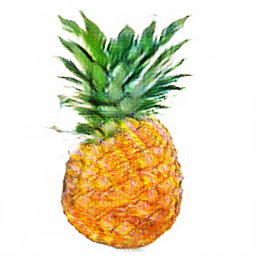}
  \end{subfigure}
  \begin{subfigure}[b]{\strwidth\linewidth}
  \includegraphics[width=\linewidth]{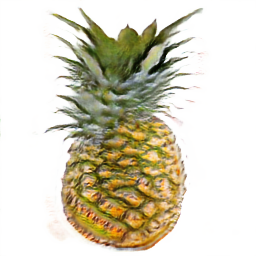}
  \end{subfigure}
  
  \begin{subfigure}[b]{\strwidth\linewidth}
  \includegraphics[width=\linewidth]{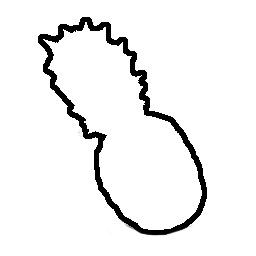}
  \end{subfigure}
   \begin{subfigure}[b]{\strwidth\linewidth}
  \includegraphics[width=\linewidth]{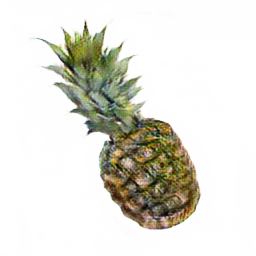}
  \end{subfigure} 
  \begin{subfigure}[b]{\strwidth\linewidth}
  \includegraphics[width=\linewidth]{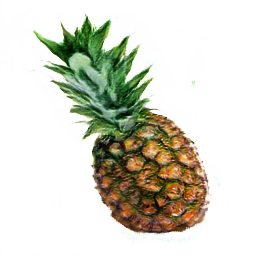}
  \end{subfigure}
   \begin{subfigure}[b]{\strwidth\linewidth}
  \includegraphics[width=\linewidth]{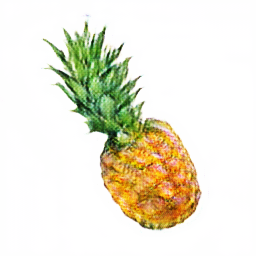}
  \end{subfigure}
  \begin{subfigure}[b]{\strwidth\linewidth}
  \includegraphics[width=\linewidth]{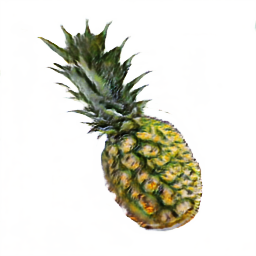}
  \end{subfigure}

  \begin{subfigure}[b]{\strwidth\linewidth}
  \includegraphics[width=\linewidth]{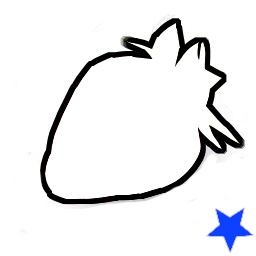}
  \end{subfigure}
  \begin{subfigure}[b]{\strwidth\linewidth}
  \includegraphics[width=\linewidth]{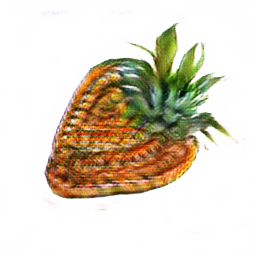}
  \end{subfigure}
  \begin{subfigure}[b]{\strwidth\linewidth}
  \includegraphics[width=\linewidth]{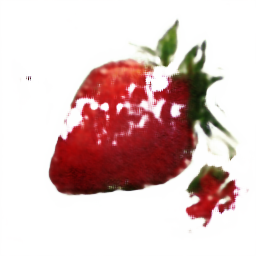}
  \end{subfigure}
  \begin{subfigure}[b]{\strwidth\linewidth}
  \includegraphics[width=\linewidth]{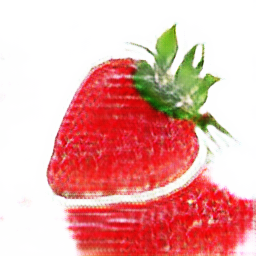}
  \end{subfigure}
  \begin{subfigure}[b]{\strwidth\linewidth}
  \includegraphics[width=\linewidth]{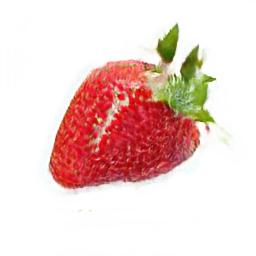}
  \end{subfigure}
  
  \begin{subfigure}[b]{\strwidth\linewidth}
  \includegraphics[width=\linewidth]{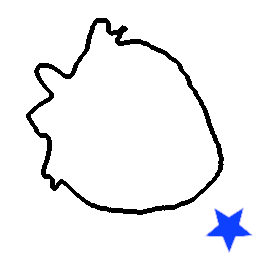}
  \subcaption{Input}
  \end{subfigure}
   \begin{subfigure}[b]{\strwidth\linewidth}
  \includegraphics[width=\linewidth]{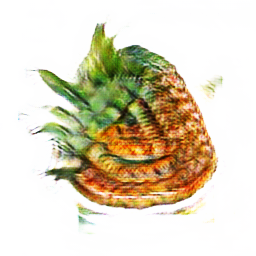}
  \caption{(a)}
  \end{subfigure} 
  \begin{subfigure}[b]{\strwidth\linewidth}
  \includegraphics[width=\linewidth]{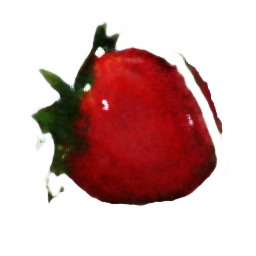}
  \subcaption{(b)}
  \end{subfigure}
   \begin{subfigure}[b]{\strwidth\linewidth}
  \includegraphics[width=\linewidth]{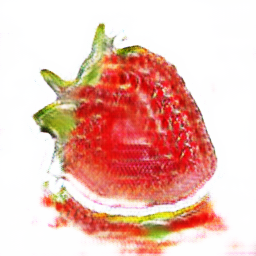}
  \subcaption{(c)}
  \end{subfigure}
  \begin{subfigure}[b]{\strwidth\linewidth}
  \includegraphics[width=\linewidth]{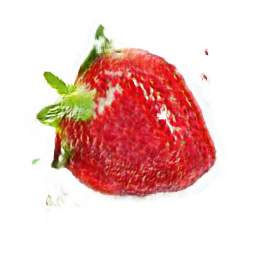}
  \subcaption{(d)}
  \end{subfigure}
\end{center}
\caption{Ablation study of the proposed solution. (a): baseline without classifier or strategy; (b): our framework without strategy; (c) trained with pre-extracted open-domain and real in-domain sketches; (d): random-mixed sampling strategy. Open-domain class is marked with \textcolor{blue}{$\bigstar$}.}
 \label{fig:abl_strategy}
\end{figure}

\begin{figure*}[htbp]
    \centering
    \includegraphics[width=.9\textwidth]{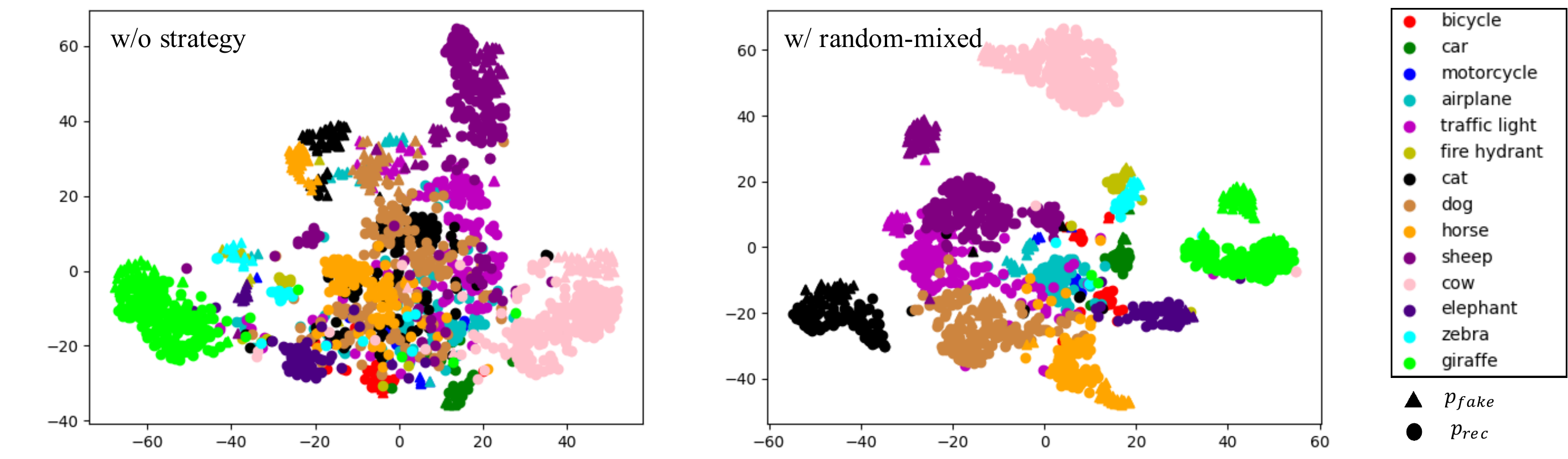}
    \caption{t-SNE visualization of photo embeddings from without any strategy, and with the random-mixed sampling strategy models. Different colors refer to different categories. Our strategies can make the generator learn more separable embeddings for different categories, regardless of in-domain or open-domain data.}
    \label{fig:tsne_strategies}
\end{figure*}

From Figure~\ref{fig:abl_strategy}, we can see that the base model in column (a) translates all inputs to the in-domain category; (b) generates texture-less images with correct colors for the open-domain class due to the pixel-wise consistency, as discussed in Equation~\ref{eq:lgan_regress}. For in-domain sketches, it generates photo-realistic outputs with the shape and texture of any category, which indicates that the model associates the class label with real/fake sketches, and thus fails to generalize to open-domain data. For column (c), the model trained with fake open-domain sketches can barely generate realistic textures for \textit{strawberries}. Besides, it fails to distinguish the object region from the background due to the weak generalization capability, as the extracted sketches actually impair the discriminative power of $D_s$. Column (d) shows that our open-domain sampling and training strategy can alleviate the above issues, and improve multi-class generation. 

To better understand the effect of the random-mixed strategy, we visualize the embedding of generated photos using the t-SNE~\cite{tsne2008laurens} on SketchyCOCO~\cite{gao2020sketchycoco}. We compare the outputs of the AODA framework trained with/without the strategy in Figure~\ref{fig:tsne_strategies}. We plot both photos $p_{fake}$ synthesized from real sketches ($\bullet$), and photos $p_{rec}$ reconstructed from fake sketches ($\blacktriangle$).As shown in the left plot, for the model trained without any strategy, even with class label conditioning, embeddings of different categories severely overlap. For most in-domain classes, the distance between $p_{fake}$ and $p_{rec}$ is much larger than the inter-class distance. At the same time, the distribution of open-domain classes (\textcolor{violet}{$\bullet$} and \textcolor{green}{$\bullet$}) is well-separated from the in-domain classes, which implies that this model cannot overcome the gap between the in-domain and open-domain data thus fails to synthesize realistic and distinct photos for multiple classes. Instead, it associates the open-domain generation's regressed objective function (Equation~\ref{eq:lgan_regress}) with the class label conditioning. As a result, the bias caused by missing sketches in the training set is amplified. While in the right plot, those issues are greatly alleviated with our proposed training strategy. The inter-class distances are greatly maximized with the aid of the random-mixed sampling strategy, which corresponds to more distinctive visual features (textures, colors, shapes, \etc) for each category. The intra-class distances are minimized, as shown in the right figure. This is likely due to the blind mixed sampling implicitly encouraging the sketch-to-image generator to ignore the domain gap between real and fake sketch inputs for all classes.



\noindent \textbf{Influence of Missing Classes} We study the influence of missing sketches by changing the number of open-domain classes $n_{\mathcal{M}}$ on the Scribble dataset. $n_{\mathcal{M}}$ increases from 0 to 6 by the following order: $strawberry$, $chicken$, $cupcake$, $moon$, $soccer$, and $basketball$. As shown in Table~\ref{tab:abl_missing}, when the number of missing classes becomes larger, the FID score increases, which means that overall output quality degrades due to the fewer real sketches in the training set. But the classification accuracy does not show such a decreasing trend thanks to the classifier's supervision. Figure~\ref{fig:abl_missing} provides visual results showing that the quality degradation exists in both in- and open-domain classes. Even so, the most representative color composition and textures of each category are maintained, making the synthesized photos recognizable for human viewers and the trained classifier.

\begin{table}[tbp]

\begin{center}
\resizebox{\columnwidth}{!}{
\begin{tabular}{l|ccccccc}
\hline
$n_{\mathcal{M}}=$                                                                       & 0 & 1 & 2   & 3 & 4 & 5 & 6 \\ \hline \hline
FID $\downarrow$        &   167.8   & 182.6  & 202.0  & 207.2 & 204.2  & 183.2 & 209.5        \\
\hline
Acc (\%) $\uparrow$  &    88.0    & 80.0   & 88.0   &  90.0 & 76.0 & 86.0 & 100.0  \\
\hline
\end{tabular}
}
\caption{Influence of missing class number on Scribble~\cite{ghosh2019interactive}.}

\label{tab:abl_missing}
\end{center}
\end{table}

\newcommand{\nmwidth}{0.11} 
\begin{figure}[tbp]
\captionsetup[subfigure]{labelformat=empty}
\begin{center}
  \begin{subfigure}[b]{\nmwidth\linewidth}
  \includegraphics[width=\linewidth]{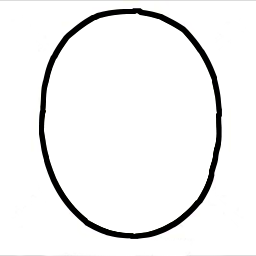}
  \end{subfigure}
  \begin{subfigure}[b]{\nmwidth\linewidth}
  \includegraphics[width=\linewidth]{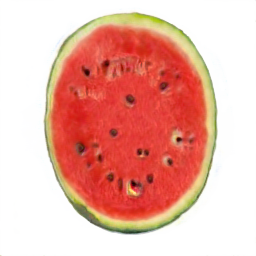}
  \end{subfigure}
\begin{subfigure}[b]{\nmwidth\linewidth}
  \includegraphics[width=\linewidth]{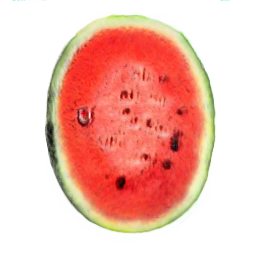}
  \end{subfigure}
  \begin{subfigure}[b]{\nmwidth\linewidth}
  \includegraphics[width=\linewidth]{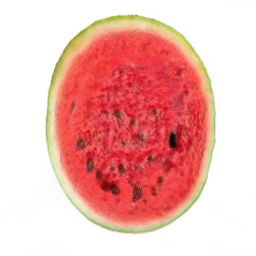}
  \end{subfigure}
  \begin{subfigure}[b]{\nmwidth\linewidth}
  \includegraphics[width=\linewidth]{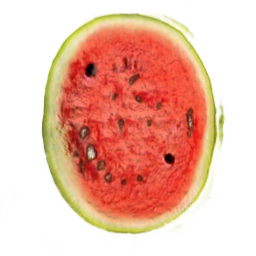}
  \end{subfigure}
  \begin{subfigure}[b]{\nmwidth\linewidth}
  \includegraphics[width=\linewidth]{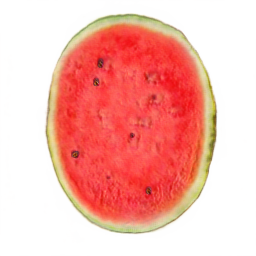}
  \end{subfigure}
\begin{subfigure}[b]{\nmwidth\linewidth}
  \includegraphics[width=\linewidth]{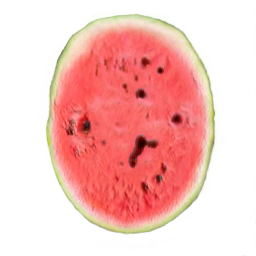}
  \end{subfigure}
  \begin{subfigure}[b]{\nmwidth\linewidth}
  \includegraphics[width=\linewidth]{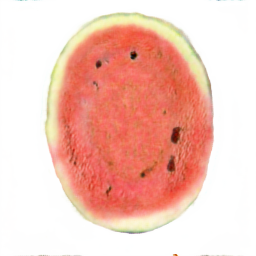}
  \end{subfigure}
  
\begin{subfigure}[b]{\nmwidth\linewidth}
  \includegraphics[width=\linewidth]{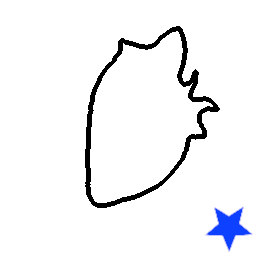}
  \subcaption{$n_{\mathcal{M}}=$ }
  \end{subfigure}
  \begin{subfigure}[b]{\nmwidth\linewidth}
  \includegraphics[width=\linewidth]{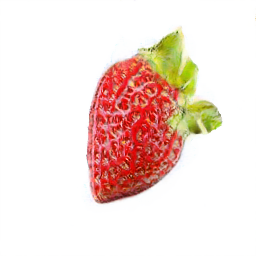}
  \subcaption{0}
  \end{subfigure}
\begin{subfigure}[b]{\nmwidth\linewidth}
  \includegraphics[width=\linewidth]{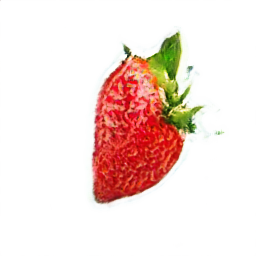}
  \subcaption{1}
  \end{subfigure}
  \begin{subfigure}[b]{\nmwidth\linewidth}
  \includegraphics[width=\linewidth]{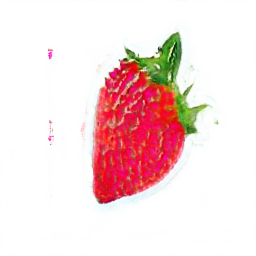}
  \subcaption{2 }
  \end{subfigure}
  \begin{subfigure}[b]{\nmwidth\linewidth}
  \includegraphics[width=\linewidth]{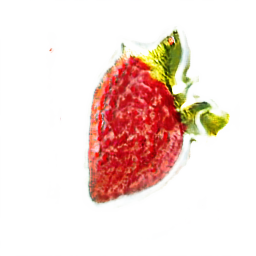}
  \subcaption{3}
  \end{subfigure}
  \begin{subfigure}[b]{\nmwidth\linewidth}
  \includegraphics[width=\linewidth]{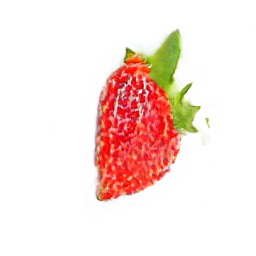}
  \subcaption{4 }
  \end{subfigure}
\begin{subfigure}[b]{\nmwidth\linewidth}
  \includegraphics[width=\linewidth]{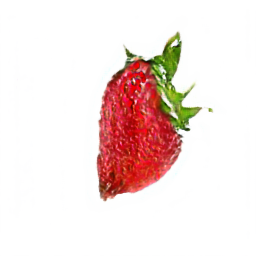}
  \subcaption{5}
  \end{subfigure}
  \begin{subfigure}[b]{\nmwidth\linewidth}
  \includegraphics[width=\linewidth]{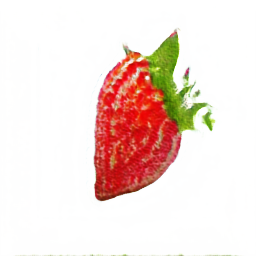}
  \subcaption{6}
  \end{subfigure}
\end{center}
\caption{Examples for the influence of missing sketches on Scribble~\cite{ghosh2019interactive}. The output quality of both in-domain and open-domain (\textcolor{blue}{$\bigstar$}) classes degrades with the increase of $n_{\mathcal{M}}$.}
 \label{fig:abl_missing}
\end{figure}

\section{Conclusion and Future Work}
In this paper, we propose an adversarial open domain adaptation framework to synthesize realistic photos from freehand sketches with class labels even if the training sketches are absent for the class. The two key ideas are that our framework (1) jointly learns sketch-to-photo and photo-to-sketch translation to make unsupervised open-domain adaptation  possible, and (2) applies the proposed open-domain training strategy to minimize the domain gap's influence on the generator and better generalize the learned correspondence of in-domain sketch-photo samples to open-domain categories. Extensive experiments and user studies on diverse datasets demonstrate that our model can faithfully synthesize realistic photos for different categories of open-domain freehand sketches. We believe that AODA provides a novel idea to utilize scarce data in real-world scenarios. In future works, we will expand our method to handle more categories of natural images and explore a more efficient design to generate higher-resolution photos. 


\clearpage
\pagebreak

\twocolumn[{
\begin{center}
  \textbf{\large Adversarial Open Domain Adaptation for Sketch-to-Photo Synthesis\\Supplementary  Material}\\[.2cm]
  Xiaoyu Xiang$^1$\footnote{The author was with Purdue University when conducting the work in this paper during an internship at ByteDance. She is now with Facebook.},  Ding Liu$^2$,  Xiao Yang$^2$, Yiheng Zhu$^2$, Xiaohui Shen$^2$, Jan P. Allebach$^1$ \\[.1cm]
  {\itshape $^{1}$Purdue University, $^{2}$ByteDance Inc.\\
  \tt\small \{xiang43,allebach\}@purdue.edu, \\
  \tt\small{\{liuding,yangxiao.0,yiheng.zhu,shenxiaohui\}@bytedance.com} \\
  }
\end{center}
\vspace{2mm}
}]




\appendix

\section{Experimental Details} 
\label{sec:exp_details}
In this section, we first illustrate the architectures of our framework, including generators, discriminators, and a classifier in Section~\ref{sec:arch}. Then, we present the objective functions for training them in Section~\ref{sec:obj}. The training settings of each dataset and additional implementation details are described in Section~\ref{sec:data} and Section~\ref{sec:imple}.

\subsection{Architecture}
\label{sec:arch}
Note that our proposed solution is not limited to certain network architecture. In this work, we select the CycleGAN~\cite{zhu2017unpaired} as a baseline to illustrate the effectiveness of our proposed solution. Thus we only modify the $G_p$ into a multi-class generator and keep the rest structures unchanged, as introduced below.

\noindent \textbf{Photo-to-Sketch Generator $G_s$}  We adopt the architecture of the photo-to-sketch generator from Johnson \etal~\cite{johnson2016perceptual}. It includes one convolution layer to map the RGB image to feature space, two downsampling layers, nine residual blocks, two upsampling layers, and one convolution layer that maps features back to the RGB image. Instance normalization~\cite{ulyanov2016instance} is used in this network. This network is also adopted as the sketch extractor for the compared method in the main paper Section 3.1.

\begin{figure}[tbp]
    \centering
    \includegraphics[width=\linewidth]{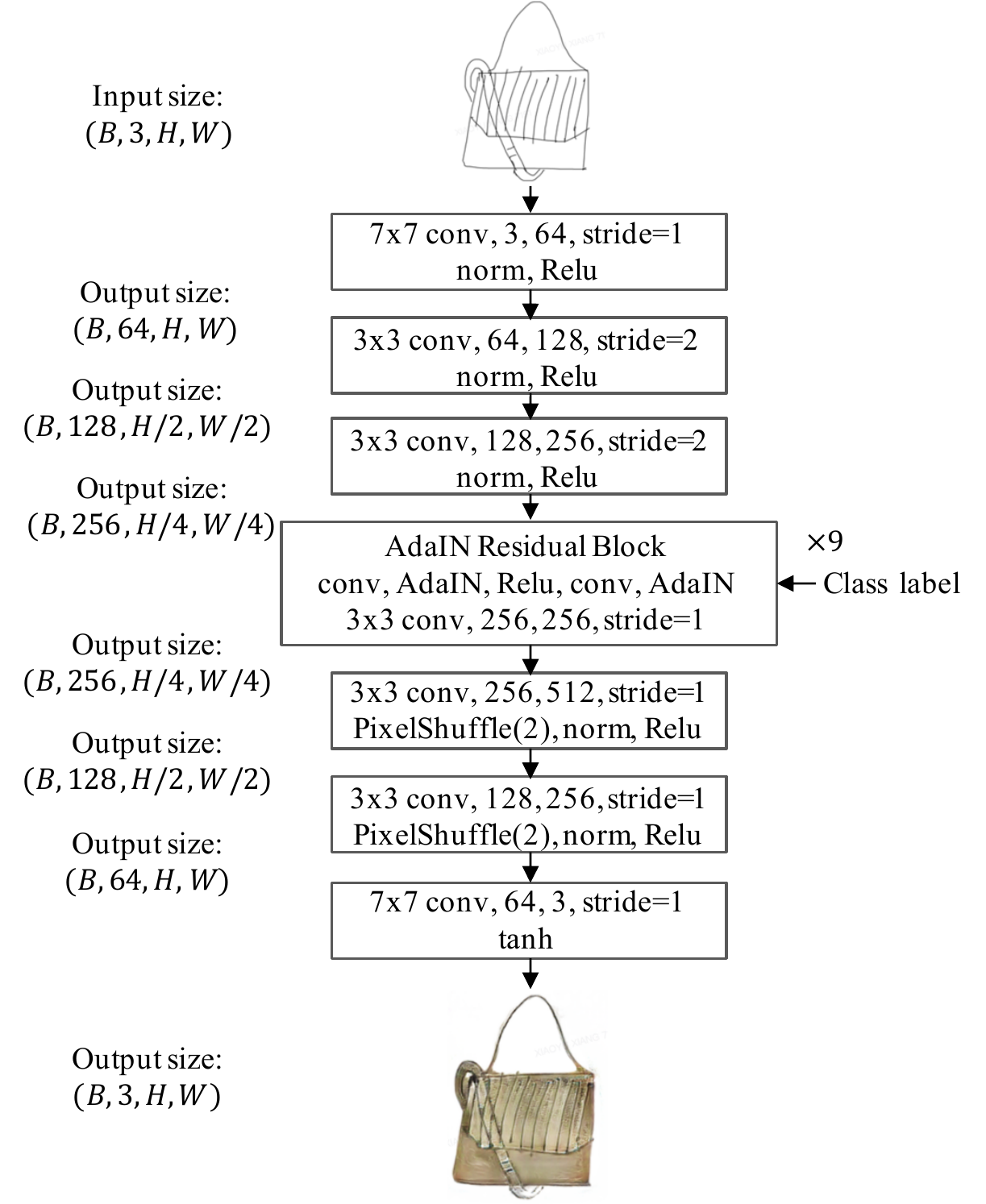}
    \caption{The architecture of our multi-class sketch-to-photo generator.} 
    \label{fig:netGB}
\end{figure}

\noindent \textbf{Multi-class Sketch-to-Photo Generator $G_p$} The overall structure of this network is similar to $G_s$: a feature-mapping convolution, two downsampling layers, a few residual blocks, two upsampling layers, and the RGB-mapping convolution. We make the following modifications on the residual blocks and upsampling layers for the multi-class photo generation, as illustrated in Figure~\ref{fig:netGB}. To make the network capable of accepting class label information, we change the normalization layers of the residual blocks into adaptive instance normalization (AdaIN)~\cite{huang2017arbitrary}. The sketch input serves as the content input for AdaIN, and the class label is the style input ensuring that the network learns the correct textures and colors for each category. In addition, we use convolution and PixelShuffle layers~\cite{shi2016real}, instead of commonly used transposed convolution, to upsample the features. The sub-pixel convolution can alleviate the checkerboard artifacts in generated photos while reducing the number of parameters as well as computations~\cite{aitken2017checkerboard}.

\noindent \textbf{Discriminators} We use the PatchGAN~\cite{li2016precomputed,ledig2017photo} classifier as the architecture for the two discriminators in our framework. It includes five convolutional layers and turns a $256\times 256$ input image into an output tensor of size $30\times 30$, where each value represents the prediction result for a $70\times 70$ patch of the input image. The final prediction output of the whole image is the average value of every patch.

\noindent \textbf{Photo Classifier} We adopt the architecture of HRNet~\cite{wang2020deep} for photo classification and change its output size of the last fully-connected (FC) layer according to the number of classes in our training data. This network takes a $256\times 256$ image as input and outputs an $n$-dim vector as the prediction result. We choose the HRNet because of its superior performance in maintaining high-resolution representations through the whole process while fusing the multi-resolution information at different stages of the network.

\subsection{Objective Function} 
\label{sec:obj}
The loss for training the generator is composed of four parts: the adversarial loss of photo-to-sketch generation $\mathcal{L}_{G_s}$, the adversarial loss of sketch-to-photo translation $\mathcal{L}_{G_p}$, the pixel-wise consistency of photo reconstruction $\mathcal{L}_{pix}$, and the classification loss for synthesized photo $\mathcal{L}_{\eta}$:
\begin{multline}
\label{eq:l_gan}
    \mathcal{L}_{GAN} = \lambda_s\mathcal{L}_{G_s}(G_s, D_s, p) + \lambda_p\mathcal{L}_{G_p}(G_p,D_p,s,\eta_s) \\ + \lambda_{pix}\mathcal{L}_{pix}(G_s,G_p,p,\eta_p) + \lambda_{\eta}\mathcal{L}_{\eta}(R,G_p,s,\eta_s),
\end{multline}

\noindent where 
\begin{equation}
    \mathcal{L}_{G_s}(G_s, D_s,p) = -\mathbb{E}_{p\sim P_{data}(p)}[\log_{D_s}(G_s(p))],
\end{equation}
\begin{equation}
    \mathcal{L}_{G_p}(G_p, D_p,s,\eta_s) = -\mathbb{E}_{s\sim P_{data}(s)}[\log_{D_p}(G_p(s, \eta_s))],
\end{equation}
\begin{equation}
    \mathcal{L}_{pix}(G_s, G_p, p, \eta_p) = \mathbb{E}_{p\sim P_{data}(p)}[||G_p(G_s(p),\eta_p)-p||_1],
\end{equation}
\begin{multline}
    \mathcal{L}_{\eta}(R,G_p,s,\eta_s) = \\ \mathbb{E}[\log P(R(G_p(s,\eta_s))=\eta_s |G_p(s,\eta_s))].
\end{multline}

Note that only the classification loss of the generated photo $G_p(s, \eta_s)$ is used to optimize the generators. 

Then we update the discriminators $D_s$ and $D_p$ with the following loss functions, respectively:
\begin{multline}
    \mathcal{L}_{D_s}(G_s, D_s,p,s) = -\mathbb{E}_{s\sim P_{data}(s)}[\log D_s(s)] \\ +\mathbb{E}_{p\sim P_{data}(p)}[\log D_s(G_s(p))],
\end{multline}
\begin{multline}
    \mathcal{L}_{D_p}(G_p, D_p, s, p,\eta_s) = -\mathbb{E}_{p\sim P_{data}(p)}[\log D_p(p)] \\ +\mathbb{E}_{s\sim P_{data}(s)}[\log D_p(G_p(s, \eta_s))].
\end{multline}

Then we calculate the classification loss of both real and synthesized photos and optimize the classifier:
\begin{multline}
    \mathcal{L}_R(R,G_p,s,p,\eta_s,\eta_p) = \mathbb{E}[\log P(R(p)=\eta_p|p)] \\ + \mathbb{E}[\log P(R(G_p (s, \eta_s))=\eta_s|G_p(s,\eta_s))].
\end{multline}

Real images and their labels enable the classifier to learn the decision boundary for each class, and the synthesized images can force the classifier to treat the fake images as the real ones and provide discriminant outputs regardless of their domain gap. For this reason, the classifier needs to be trained jointly with the other parts of our framework.

We adopt the binary cross-entropy loss for discriminators and focal loss~\cite{lin2017focal} for classification. The pixel-wise loss for photo reconstruction is measured by $\mathcal{L}1$-distance.

\subsection{Datasets}
\label{sec:data}
We train our model on three datasets: Scribble~\cite{ghosh2019interactive} (10 classes), QMUL-Sketch~\cite{yu2016sketch,song2017deep,liu2019unpaired} (3 classes), and SketchyCOCO~\cite{gao2020sketchycoco} (14 classes of objects). During the training stage, the sketches of certain classes are completely removed to meet the open-domain settings.

\noindent \textbf{Scribble} This dataset contains ten classes of objects, including white-background photos and simple outline sketches. Six out of ten object classes have similar round outlines, which imposes more stringent requirements on the network: whether it can generate the correct structure and texture conditioned on the input class label. In our open-domain setting, we only have the sketches of four classes for training: $pineapple$ (151 images), $cookie$ (147 images), $orange$ (146 images), and $watermelon$ (146 images). We set the input image size to $256\times 256$ and train all the compared networks for 200 epochs. We apply the Adam~\cite{kingma2014adam} optimizer with batch size$=1$, and the learning rate is set to $2e-4$ for the first 100 epochs, and it decreases linearly to zero in the second 100 epochs. 

\noindent \textbf{QMUL-Sketch} We construct it by combing three datasets: handbags~\cite{song2017deep} with 400 photos and sketches, ShoeV2~\cite{yu2016sketch} with 2000 photos and 6648 sketches, and ChairV2~\cite{liu2019unpaired} with 400 photos and 1297 sketches. For the open-domain training setting, we completely remove the sketches of the ChairV2. We train the networks for 400 epochs.

\noindent \textbf{SketchyCOCO} This dataset includes 14 object classes, where the sketches are collected from the Sketchy dataset~\cite{sangkloy2016sketchy}, TU-Berlin dataset~\cite{eitz2012humans}, and $Quick! Draw$
dataset~\cite{ha2017neural}. The 14,081 photos for each object class are segmented from the natural images of COCO Stuff~\cite{caesar2018coco} under unconstrained conditions, thereby making it more difficult for existing methods to map the freehand sketches to the photo domain. In our open-domain setting, we remove the sketches of two classes during training: $sheep$ and $giraffe$. We use EdgeGAN weights released by the author. All the other networks are trained for 100 epochs.

\subsection{Implementation Details} 
\label{sec:imple}
Our model is implemented in PyTorch~\cite{paszke2017automatic,paszke2019pytorch}. We train our networks with the standard Adam~\cite{kingma2014adam} using 1 NVIDIA V100 GPU. The batch size and initial learning rate are set to $1$ and $2e-4$ for all datasets. The epoch numbers are 200, 400, and 100 for the Scribble~\cite{ghosh2019interactive}, QMUL-Sketch~\cite{yu2016sketch,song2017deep,liu2019unpaired}, and SketchyCOCO~\cite{gao2020sketchycoco}, respectively. The learning rates drop by multiplying $0.5$ in the second half of epochs. 
For the compared method EdgeGAN~\cite{gao2020sketchycoco}, we use the official implementation in \href{https://github.com/sysu-imsl/EdgeGAN}{https://github.com/sysu-imsl/EdgeGAN} for data preprocessing and training. It is trained for 100 epochs on Scribble and QMUL datasets using one NVIDIA GTX 2080 GPU. The batch size is set to 1 due to memory limitation. 

\section{Experimental Results}
\label{sec:exp_res}
We first show more sketch-to-photo results on the QMUL-Sketch dataset in Section~\ref{sec:qmul} and briefly discuss these results. At last, we show more sketch-to-photo synthesis results of our method in Section~\ref{sec:s2p_syn}. 

\subsection{Comparison on QMUL-Sketch Dataset}
\label{sec:qmul}
 We compare our method with the same baseline methods as described in the main paper: (a) CycleGAN as the baseline, (b) conditional CycleGAN that takes sketch and class label as input, and (c) EdgeGAN~\cite{gao2020sketchycoco} trained on this dataset. Different from the Scribble dataset, the sketches in QMUL-Sketch are from three different datasets with rich strokes. Thus, the sketch itself already contains sufficient class information~\cite{liu2019unpaired}. As shown in Figure~\ref{fig:qmul_results}, most compared methods can generate high-quality photos. Still, all of these methods change the structure of the open-domain class ($Chair$), as shown in the bottom two rows of columns (a), (b), and (c) of Figure~\ref{fig:qmul_results}. Compared with them, our model can maintain the natural shape in the original sketch and generate realistic photos.

The quantitative results are shown in Table~\ref{tab:qmul}. We can see that our model is preferred by more users than the other compared methods. While in terms of the FID score and classification accuracy, ours is the second-best. This is because the sketches in the QMUL-Sketch dataset are three times more than the photos (especially for \textit{shoes}), which is not consistent with our motivation of enriching the missing sketches with abundant photos. Under this scenario, the asymmetry within the framework and strategies' design does not bring too many benefits. 

\newcommand{\qmulwidth}{0.18} 
\begin{figure}[htbp]
\captionsetup[subfigure]{labelformat=empty}
\begin{center}
  \begin{subfigure}[b]{\qmulwidth\linewidth}
  \includegraphics[width=\linewidth]{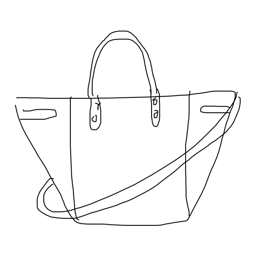}
  \end{subfigure}
  \begin{subfigure}[b]{\qmulwidth\linewidth}
  \includegraphics[width=\linewidth]{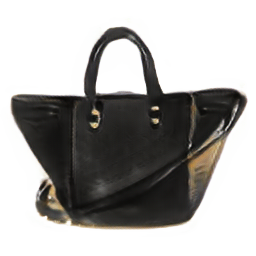}
  \end{subfigure}
  \begin{subfigure}[b]{\qmulwidth\linewidth}
  \includegraphics[width=\linewidth]{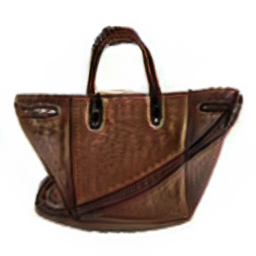}
  \end{subfigure}
   \begin{subfigure}[b]{\qmulwidth\linewidth}
  \includegraphics[width=\linewidth]{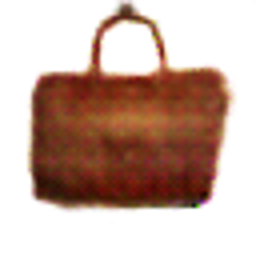}
  \end{subfigure}
  \begin{subfigure}[b]{\qmulwidth\linewidth}
  \includegraphics[width=\linewidth]{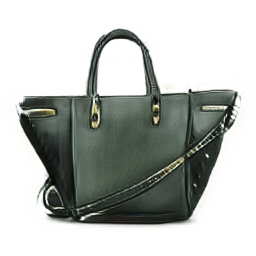}
  \end{subfigure}
  
\begin{subfigure}[b]{\qmulwidth\linewidth}
  \includegraphics[width=\linewidth]{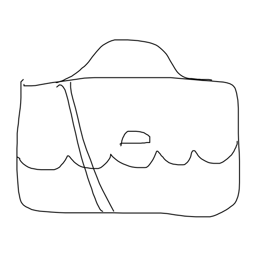}
  \end{subfigure}
  \begin{subfigure}[b]{\qmulwidth\linewidth}
  \includegraphics[width=\linewidth]{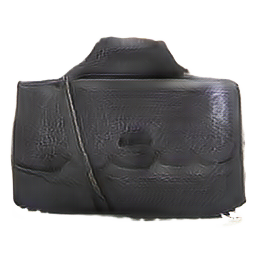}
  \end{subfigure}
  \begin{subfigure}[b]{\qmulwidth\linewidth}
  \includegraphics[width=\linewidth]{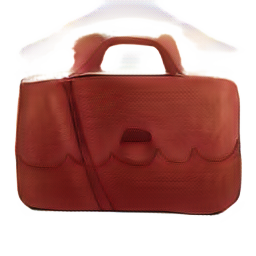}
  \end{subfigure} 
   \begin{subfigure}[b]{\qmulwidth\linewidth}
  \includegraphics[width=\linewidth]{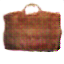}
  \end{subfigure}
  \begin{subfigure}[b]{\qmulwidth\linewidth}
  \includegraphics[width=\linewidth]{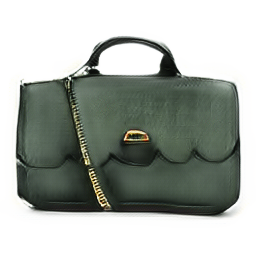}
  \end{subfigure}
  
  \begin{subfigure}[b]{\qmulwidth\linewidth}
  \includegraphics[width=\linewidth]{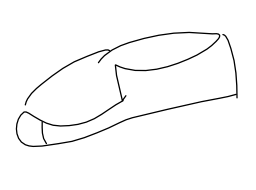}
  \end{subfigure}
  \begin{subfigure}[b]{\qmulwidth\linewidth}
  \includegraphics[width=\linewidth]{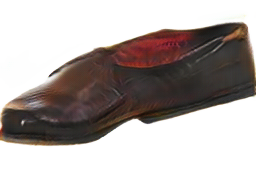}
  \end{subfigure}
   \begin{subfigure}[b]{\qmulwidth\linewidth}
  \includegraphics[width=\linewidth]{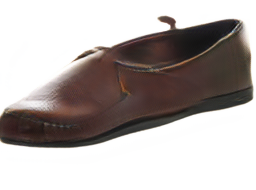}
  \end{subfigure}
   \begin{subfigure}[b]{\qmulwidth\linewidth}
  \includegraphics[width=\linewidth]{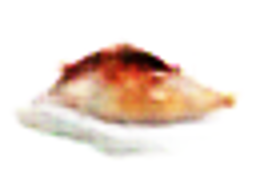}
  \end{subfigure}
  \begin{subfigure}[b]{\qmulwidth\linewidth}
  \includegraphics[width=\linewidth]{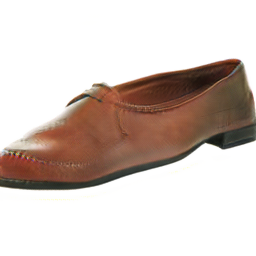}
  \end{subfigure}
  
 \begin{subfigure}[b]{\qmulwidth\linewidth}
  \includegraphics[width=\linewidth]{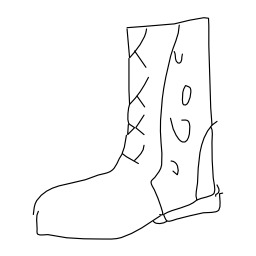}
  \end{subfigure}
  \begin{subfigure}[b]{\qmulwidth\linewidth}
  \includegraphics[width=\linewidth]{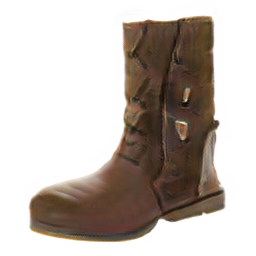}
  \end{subfigure}
  \begin{subfigure}[b]{\qmulwidth\linewidth}
  \includegraphics[width=\linewidth]{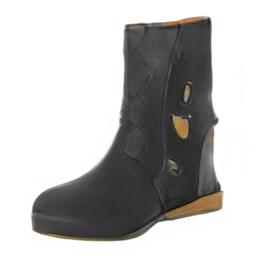}
  \end{subfigure}
   \begin{subfigure}[b]{\qmulwidth\linewidth}
  \includegraphics[width=\linewidth]{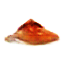}
  \end{subfigure}
  \begin{subfigure}[b]{\qmulwidth\linewidth}
  \includegraphics[width=\linewidth]{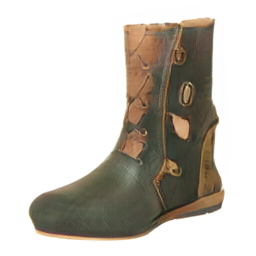}
  \end{subfigure}
  
\begin{subfigure}[b]{\qmulwidth\linewidth}
  \includegraphics[width=\linewidth]{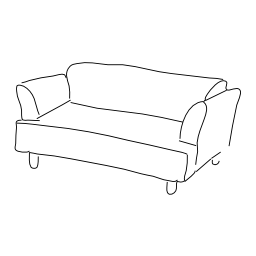}
  \end{subfigure}
  \begin{subfigure}[b]{\qmulwidth\linewidth}
  \includegraphics[width=\linewidth]{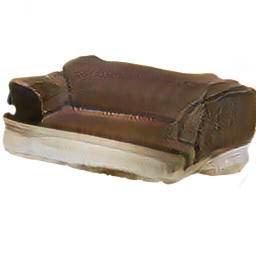}
  \end{subfigure}
   \begin{subfigure}[b]{\qmulwidth\linewidth}
  \includegraphics[width=\linewidth]{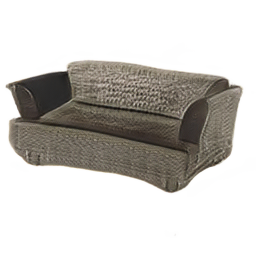}
  \end{subfigure}
   \begin{subfigure}[b]{\qmulwidth\linewidth}
  \includegraphics[width=\linewidth]{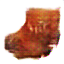}
  \end{subfigure}
  \begin{subfigure}[b]{\qmulwidth\linewidth}
  \includegraphics[width=\linewidth]{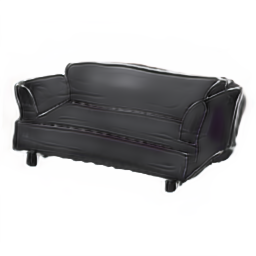}
  \end{subfigure}
  
  \begin{subfigure}[b]{\qmulwidth\linewidth}
  \includegraphics[width=\linewidth]{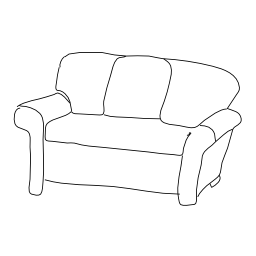}
  \subcaption{Input}
  \end{subfigure}
  \begin{subfigure}[b]{\qmulwidth\linewidth}
  \includegraphics[width=\linewidth]{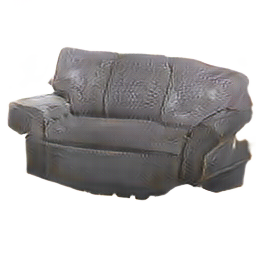}
  \subcaption{(a)}
  \end{subfigure}
  \begin{subfigure}[b]{\qmulwidth\linewidth}
  \includegraphics[width=\linewidth]{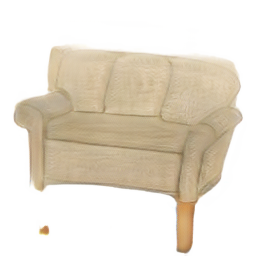}
  \subcaption{(b)}
  \end{subfigure}
\begin{subfigure}[b]{\qmulwidth\linewidth}
  \includegraphics[width=\linewidth]{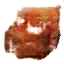}
  \subcaption{(c)}
  \end{subfigure}
  \begin{subfigure}[b]{\qmulwidth\linewidth}
  \includegraphics[width=\linewidth]{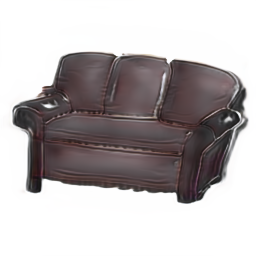}
  \subcaption{Ours}
  \end{subfigure}
\end{center}
\caption{Results on the QMUL-Sketch dataset. Compared with methods (a) CycleGAN~\cite{zhu2017unpaired}, (b) conditional CycleGAN, and (c) EdgeGAN~\cite{gao2020sketchycoco}, our model can faithfully maintain the natural shapes in sketch inputs and synthesize realistic photos.}
 \label{fig:qmul_results}
\end{figure}

\begin{table}[t]
\centering
\resizebox{\linewidth}{!}{
\begin{tabular}{c|c|ccc}
\hline
Metric     &  Method      &     full & in-domain & open-domain       \\
\hline
FID $\downarrow$         &  CycleGAN     & 97.9                     & 87.7                     & 151.7         \\
 & conditional CycleGAN & \textbf{91.6}                     & 88.2                     & \textbf{107.5}       \\
& EdgeGAN & 243.0                      & 281.3                    & 268.3           \\
& Ours & 92.4                   &   \textbf{76.9}                       &        142.6                    \\
\hline
Acc (\%) $\uparrow$ & CycleGAN &  72.6                     & 64.7                     & 78.2     \\
 & conditioanl CycleGAN & 78.4                     & 58.0                       & \textbf{92.6}            \\
 & EdgeGAN & 62.7                     & \textbf{100.0}                      & 36.8        \\
 & Ours &    \textbf{89.7}                       &         91.5                  &          88.5                  \\
 \hline 
Human (\%) $\uparrow$&  CycleGAN &  4.00   & 4.57    & 2.67    \\
& conditional CycleGAN &  21.20   & 18.86   &   26.67  \\ 
 & EdgeGAN &   0.00  &  0.00  &  0.00    \\
 & Ours &  \textbf{74.80}                         &      \textbf{76.57}                     &    \textbf{70.67}                        \\
\hline
\end{tabular}
}
\caption{Results of quantitative evaluation and user preference study on QMUL-Sketch dataset. Best results are shown in \textbf{bold}.}
\label{tab:qmul}
\end{table}


\subsection{More Sketch-to-Photo Results}
\label{sec:s2p_syn}
Here we show more $256\times 256$ sketch-to-photo results of our model in Figure~\ref{fig:hr_coco}, \ref{fig:hr_qmul} and \ref{fig:hr_scribble}. Previous sketch-to-photo synthesis works usually have output sizes $=64 \times 64$ or $128 \times 128$. Leveraging the output size makes the problem even more challenging for two reasons: (1) the difficulty of correcting larger shape deformation, and (2) generating richer details and realistic textures for each image composition. The results in the following pages suggest that AODA is able to synthesize $256\times 256$ photo-realistic images.

In addition, Figure~\ref{fig:in_res} shows the in-domain results obtained on the full dataset of Scribble~\cite{ghosh2019interactive} without removing any sketch. Our network not only handle the open-domain training problem, but also perform even better under a common multi-class sketch-to-photo generation setting.

\newcommand{\inwidth}{0.20} 
\begin{figure*}[tbp]
\captionsetup[subfigure]{labelformat=empty}
\begin{center}
\begin{subfigure}[b]{\inwidth\linewidth}
  \includegraphics[width=\linewidth]{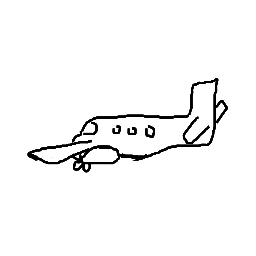}
  \end{subfigure}
  \begin{subfigure}[b]{\inwidth\linewidth}
  \includegraphics[width=\linewidth]{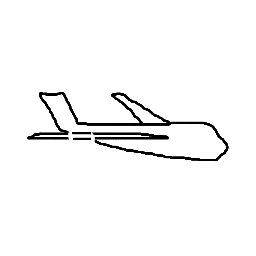}
  \end{subfigure}
  \begin{subfigure}[b]{\inwidth\linewidth}
  \includegraphics[width=\linewidth]{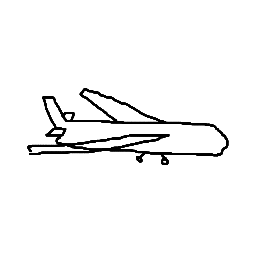}
  \end{subfigure}
\begin{subfigure}[b]{\inwidth\linewidth}
  \includegraphics[width=\linewidth]{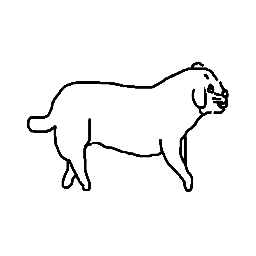}
  \end{subfigure}
  
\begin{subfigure}[b]{\inwidth\linewidth}
  \includegraphics[width=\linewidth]{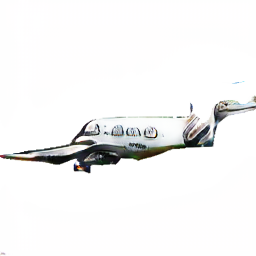}
  \end{subfigure}
  \begin{subfigure}[b]{\inwidth\linewidth}
  \includegraphics[width=\linewidth]{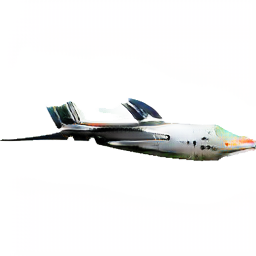}
  \end{subfigure}
  \begin{subfigure}[b]{\inwidth\linewidth}
  \includegraphics[width=\linewidth]{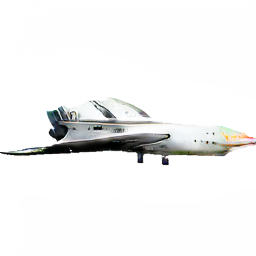}
  \end{subfigure}
\begin{subfigure}[b]{\inwidth\linewidth}
  \includegraphics[width=\linewidth]{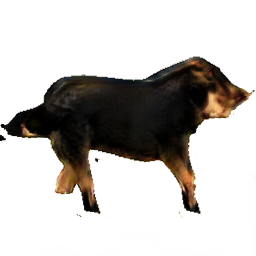}
  \end{subfigure}

  \begin{subfigure}[b]{\inwidth\linewidth}
  \includegraphics[width=\linewidth]{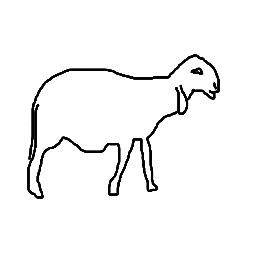}
  \end{subfigure}
  \begin{subfigure}[b]{\inwidth\linewidth}
  \includegraphics[width=\linewidth]{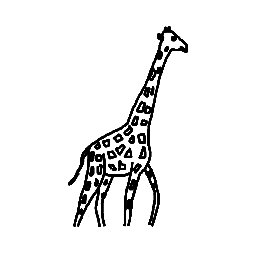}
  \end{subfigure}
\begin{subfigure}[b]{\inwidth\linewidth}
  \includegraphics[width=\linewidth]{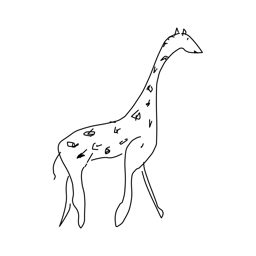}
  \end{subfigure}
  \begin{subfigure}[b]{\inwidth\linewidth}
  \includegraphics[width=\linewidth]{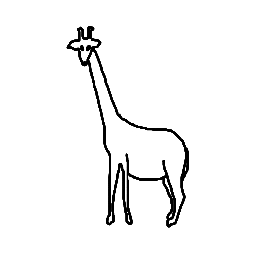}
  \end{subfigure}
  
  \begin{subfigure}[b]{\inwidth\linewidth}
  \includegraphics[width=\linewidth]{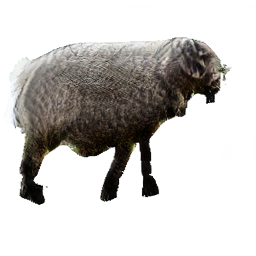}
  \end{subfigure}
  \begin{subfigure}[b]{\inwidth\linewidth}
  \includegraphics[width=\linewidth]{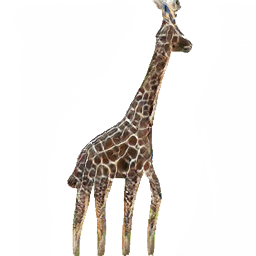}
  \end{subfigure}
\begin{subfigure}[b]{\inwidth\linewidth}
  \includegraphics[width=\linewidth]{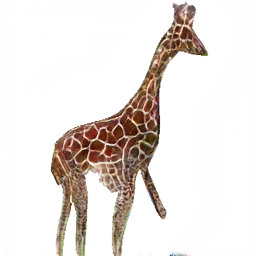}
  \end{subfigure}
  \begin{subfigure}[b]{\inwidth\linewidth}
  \includegraphics[width=\linewidth]{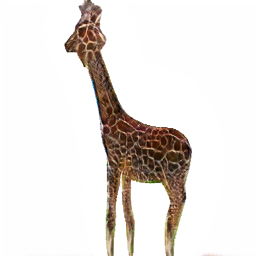}
  \end{subfigure}
  
\end{center}
\caption{More $256\times 256$ results on the SketchyCOCO dataset.}
 \label{fig:hr_coco}

\end{figure*}

\newcommand{\hrwidth}{0.20} 
\begin{figure*}[tbp]
\captionsetup[subfigure]{labelformat=empty}
\begin{center}
\begin{subfigure}[b]{\hrwidth\linewidth}
  \includegraphics[width=\linewidth]{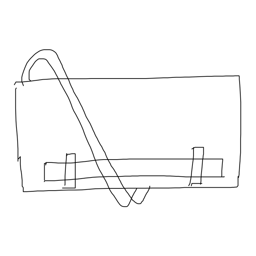}
  \end{subfigure}
  \begin{subfigure}[b]{\hrwidth\linewidth}
  \includegraphics[width=\linewidth]{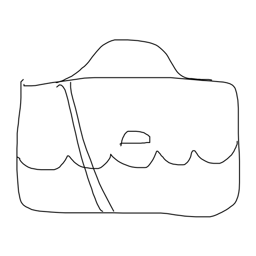}
  \end{subfigure}
\begin{subfigure}[b]{\hrwidth\linewidth}
  \includegraphics[width=\linewidth]{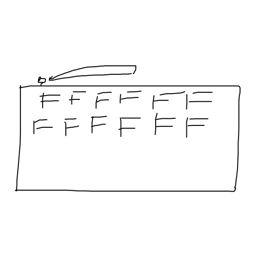}
  \end{subfigure}
  \begin{subfigure}[b]{\hrwidth\linewidth}
  \includegraphics[width=\linewidth]{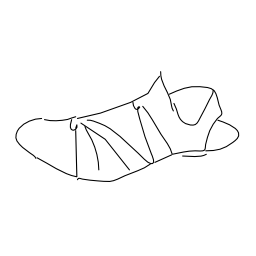}
  \end{subfigure}

\begin{subfigure}[b]{\hrwidth\linewidth}
  \includegraphics[width=\linewidth]{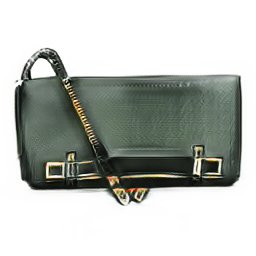}
  \end{subfigure}
  \begin{subfigure}[b]{\hrwidth\linewidth}
  \includegraphics[width=\linewidth]{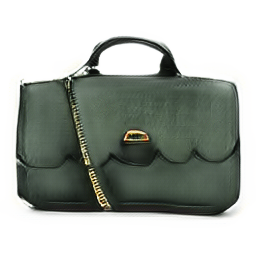}
  \end{subfigure}
\begin{subfigure}[b]{\hrwidth\linewidth}
  \includegraphics[width=\linewidth]{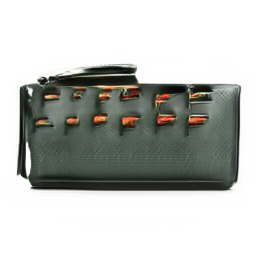}
  \end{subfigure}
  \begin{subfigure}[b]{\hrwidth\linewidth}
  \includegraphics[width=\linewidth]{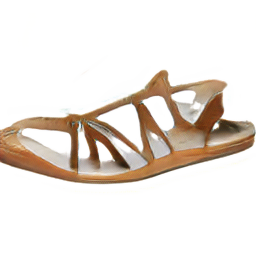}
  \end{subfigure}

\begin{subfigure}[b]{\hrwidth\linewidth}
  \includegraphics[width=\linewidth]{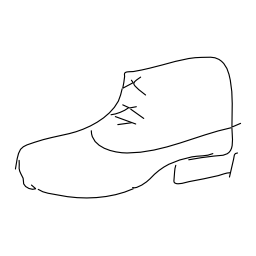}
  \end{subfigure}
  \begin{subfigure}[b]{\hrwidth\linewidth}
  \includegraphics[width=\linewidth]{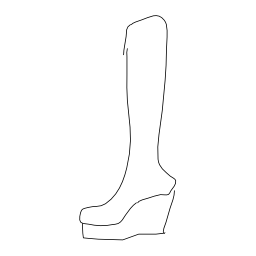}
  \end{subfigure}
\begin{subfigure}[b]{\hrwidth\linewidth}
  \includegraphics[width=\linewidth]{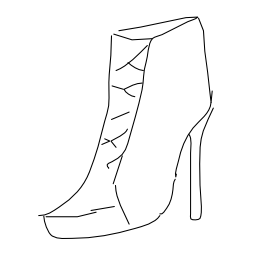}
  \end{subfigure}
  \begin{subfigure}[b]{\hrwidth\linewidth}
  \includegraphics[width=\linewidth]{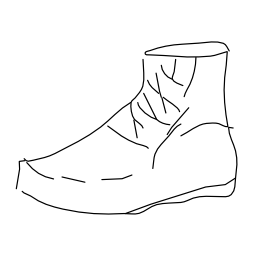}
  \end{subfigure} 

\begin{subfigure}[b]{\hrwidth\linewidth}
  \includegraphics[width=\linewidth]{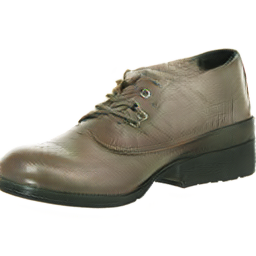}
  \end{subfigure}
  \begin{subfigure}[b]{\hrwidth\linewidth}
  \includegraphics[width=\linewidth]{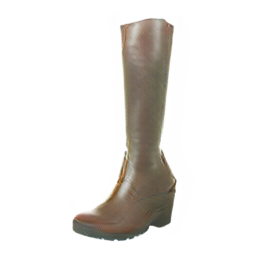}
  \end{subfigure}
\begin{subfigure}[b]{\hrwidth\linewidth}
  \includegraphics[width=\linewidth]{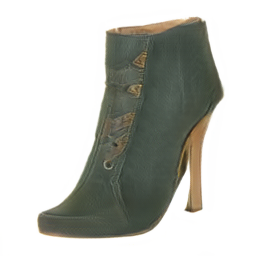}
  \end{subfigure}
  \begin{subfigure}[b]{\hrwidth\linewidth}
  \includegraphics[width=\linewidth]{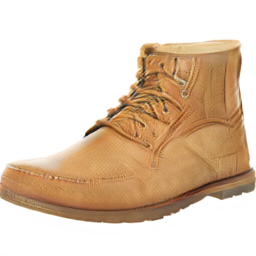}
  \end{subfigure}
  
\begin{subfigure}[b]{\hrwidth\linewidth}
  \includegraphics[width=\linewidth]{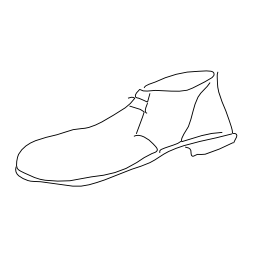}
  \end{subfigure}
  \begin{subfigure}[b]{\hrwidth\linewidth}
  \includegraphics[width=\linewidth]{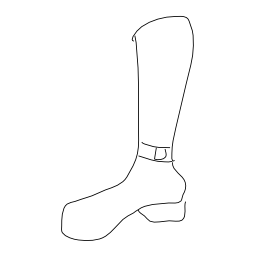}
  \end{subfigure}
\begin{subfigure}[b]{\hrwidth\linewidth}
  \includegraphics[width=\linewidth]{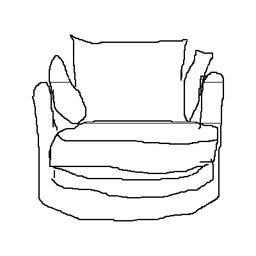}
  \end{subfigure}
  \begin{subfigure}[b]{\hrwidth\linewidth}
  \includegraphics[width=\linewidth]{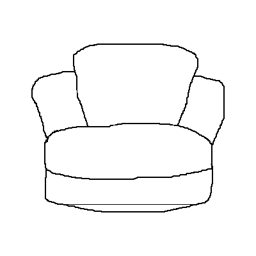}
  \end{subfigure}
  
\begin{subfigure}[b]{\hrwidth\linewidth}
  \includegraphics[width=\linewidth]{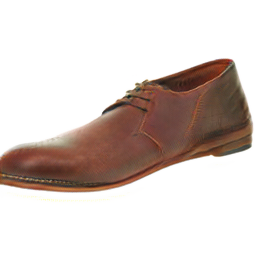}
  \end{subfigure}
  \begin{subfigure}[b]{\hrwidth\linewidth}
  \includegraphics[width=\linewidth]{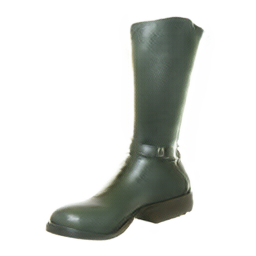}
  \end{subfigure}
\begin{subfigure}[b]{\hrwidth\linewidth}
  \includegraphics[width=\linewidth]{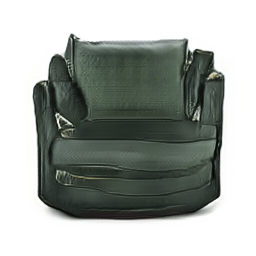}
  \end{subfigure}
  \begin{subfigure}[b]{\hrwidth\linewidth}
  \includegraphics[width=\linewidth]{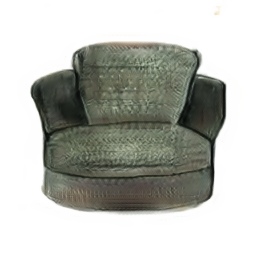}
  \end{subfigure}
  
\end{center}
\caption{More $256\times 256$ results on the QMUL-Sketch dataset.}
 \label{fig:hr_qmul}
\end{figure*}

\begin{figure*}[tbp]
\captionsetup[subfigure]{labelformat=empty}
\begin{center}
  \begin{subfigure}[b]{\hrwidth\linewidth}
  \includegraphics[width=\linewidth]{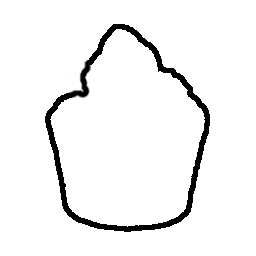}
  \end{subfigure}
  \begin{subfigure}[b]{\hrwidth\linewidth}
  \includegraphics[width=\linewidth]{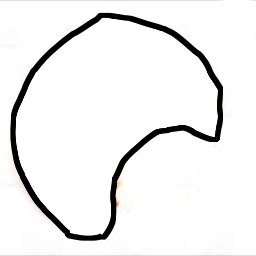}
  \end{subfigure}
\begin{subfigure}[b]{\hrwidth\linewidth}
  \includegraphics[width=\linewidth]{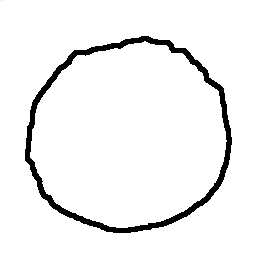}
  \end{subfigure}
  \begin{subfigure}[b]{\hrwidth\linewidth}
  \includegraphics[width=\linewidth]{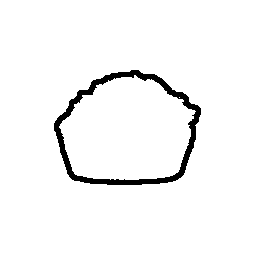}
  \end{subfigure}

  \begin{subfigure}[b]{\hrwidth\linewidth}
  \includegraphics[width=\linewidth]{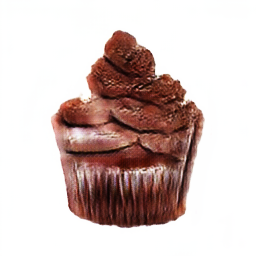}
  \end{subfigure}
  \begin{subfigure}[b]{\hrwidth\linewidth}
  \includegraphics[width=\linewidth]{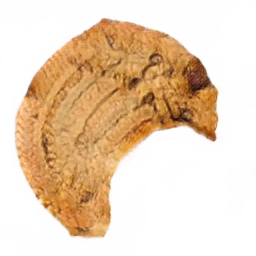}
  \end{subfigure}
\begin{subfigure}[b]{\hrwidth\linewidth}
  \includegraphics[width=\linewidth]{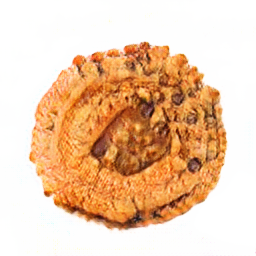}
  \end{subfigure}
  \begin{subfigure}[b]{\hrwidth\linewidth}
  \includegraphics[width=\linewidth]{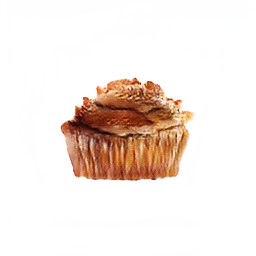}
  \end{subfigure}

\begin{subfigure}[b]{\hrwidth\linewidth}
  \includegraphics[width=\linewidth]{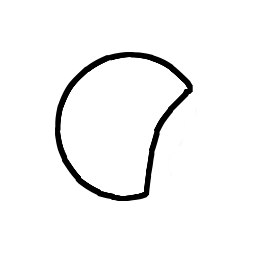}
  \end{subfigure}
  \begin{subfigure}[b]{\hrwidth\linewidth}
  \includegraphics[width=\linewidth]{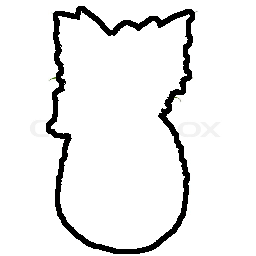}
  \end{subfigure}
\begin{subfigure}[b]{\hrwidth\linewidth}
  \includegraphics[width=\linewidth]{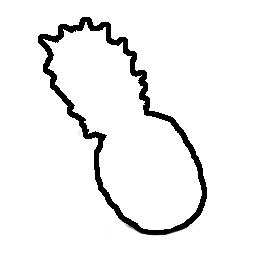}
  \end{subfigure}
  \begin{subfigure}[b]{\hrwidth\linewidth}
  \includegraphics[width=\linewidth]{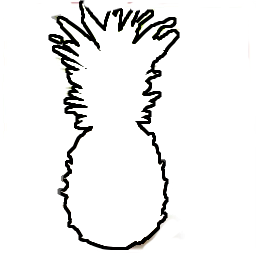}
  \end{subfigure}  

\begin{subfigure}[b]{\hrwidth\linewidth}
  \includegraphics[width=\linewidth]{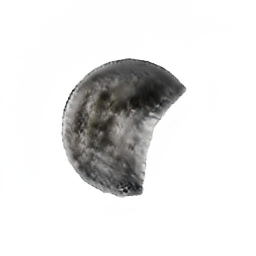}
  \end{subfigure}
  \begin{subfigure}[b]{\hrwidth\linewidth}
  \includegraphics[width=\linewidth]{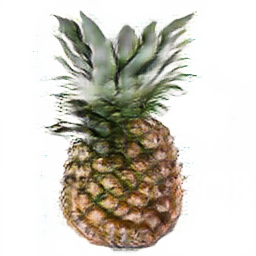}
  \end{subfigure}
\begin{subfigure}[b]{\hrwidth\linewidth}
  \includegraphics[width=\linewidth]{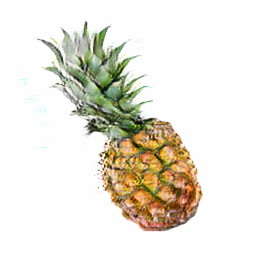}
  \end{subfigure}
  \begin{subfigure}[b]{\hrwidth\linewidth}
  \includegraphics[width=\linewidth]{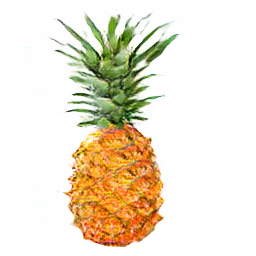}
  \end{subfigure}
  
\begin{subfigure}[b]{\hrwidth\linewidth}
  \includegraphics[width=\linewidth]{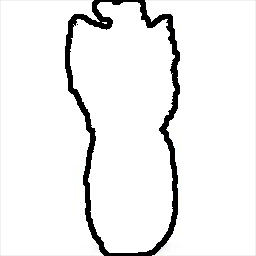}
  \end{subfigure}
  \begin{subfigure}[b]{\hrwidth\linewidth}
  \includegraphics[width=\linewidth]{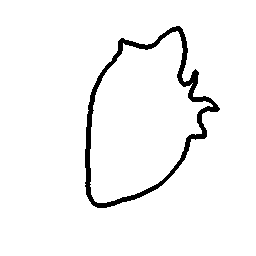}
  \end{subfigure}
\begin{subfigure}[b]{\hrwidth\linewidth}
  \includegraphics[width=\linewidth]{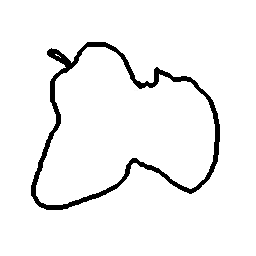}
  \end{subfigure}
  \begin{subfigure}[b]{\hrwidth\linewidth}
  \includegraphics[width=\linewidth]{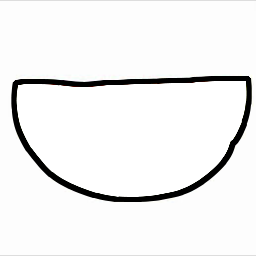}
  \end{subfigure}
  
\begin{subfigure}[b]{\hrwidth\linewidth}
  \includegraphics[width=\linewidth]{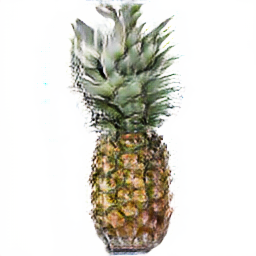}
  \end{subfigure}
  \begin{subfigure}[b]{\hrwidth\linewidth}
  \includegraphics[width=\linewidth]{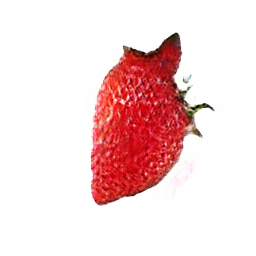}
  \end{subfigure}
\begin{subfigure}[b]{\hrwidth\linewidth}
  \includegraphics[width=\linewidth]{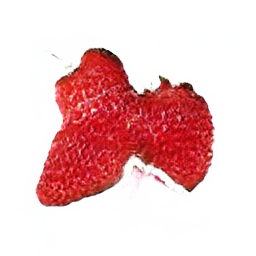}
  \end{subfigure}
  \begin{subfigure}[b]{\hrwidth\linewidth}
  \includegraphics[width=\linewidth]{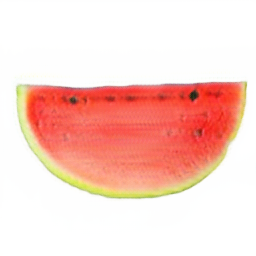}
  \end{subfigure}
  
\end{center}
\caption{More $256\times 256$ results on the Scribble dataset.}
 \label{fig:hr_scribble}
\end{figure*}

\begin{figure*}[tbp]
\captionsetup[subfigure]{labelformat=empty}
\begin{center}
  \begin{subfigure}[b]{\hrwidth\linewidth}
  \includegraphics[width=\linewidth]{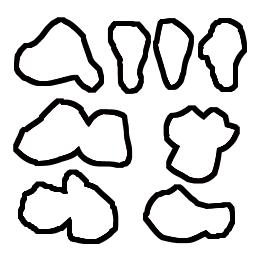}
  \end{subfigure}
  \begin{subfigure}[b]{\hrwidth\linewidth}
  \includegraphics[width=\linewidth]{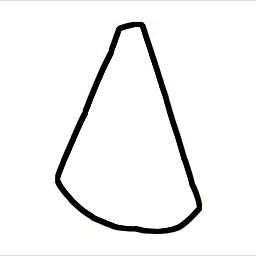}
  \end{subfigure}
\begin{subfigure}[b]{\hrwidth\linewidth}
  \includegraphics[width=\linewidth]{img/hr/in/cookie_147_real_B.png}
  \end{subfigure}
  \begin{subfigure}[b]{\hrwidth\linewidth}
  \includegraphics[width=\linewidth]{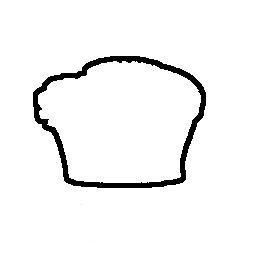}
  \end{subfigure}

  \begin{subfigure}[b]{\hrwidth\linewidth}
  \includegraphics[width=\linewidth]{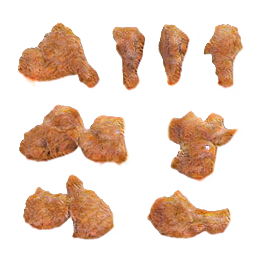}
  \end{subfigure}
  \begin{subfigure}[b]{\hrwidth\linewidth}
  \includegraphics[width=\linewidth]{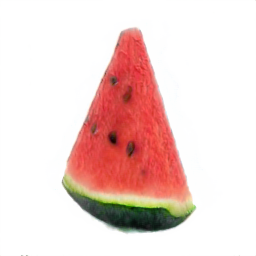}
  \end{subfigure}
\begin{subfigure}[b]{\hrwidth\linewidth}
  \includegraphics[width=\linewidth]{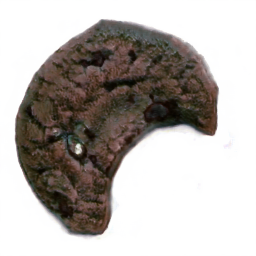}
  \end{subfigure}
  \begin{subfigure}[b]{\hrwidth\linewidth}
  \includegraphics[width=\linewidth]{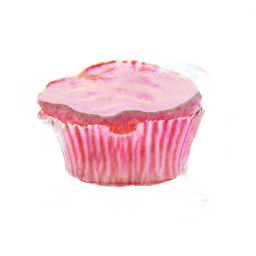}
  \end{subfigure}

  \begin{subfigure}[b]{\hrwidth\linewidth}
  \includegraphics[width=\linewidth]{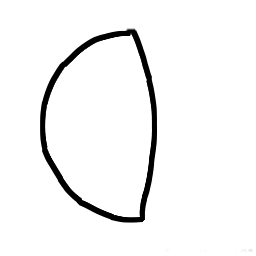}
  \end{subfigure}
  \begin{subfigure}[b]{\hrwidth\linewidth}
  \includegraphics[width=\linewidth]{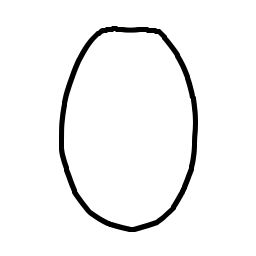}
  \end{subfigure}
\begin{subfigure}[b]{\hrwidth\linewidth}
  \includegraphics[width=\linewidth]{img/hr/in/pineapple_154_real_B.png}
  \end{subfigure}
  \begin{subfigure}[b]{\hrwidth\linewidth}
  \includegraphics[width=\linewidth]{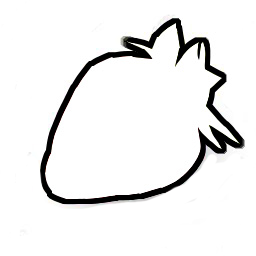}
  \end{subfigure}

  \begin{subfigure}[b]{\hrwidth\linewidth}
  \includegraphics[width=\linewidth]{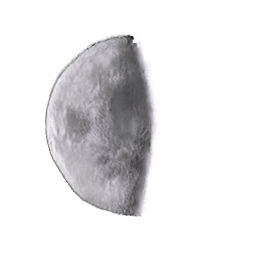}
  \end{subfigure}
  \begin{subfigure}[b]{\hrwidth\linewidth}
  \includegraphics[width=\linewidth]{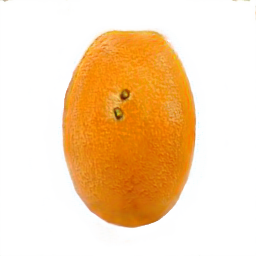}
  \end{subfigure}
\begin{subfigure}[b]{\hrwidth\linewidth}
  \includegraphics[width=\linewidth]{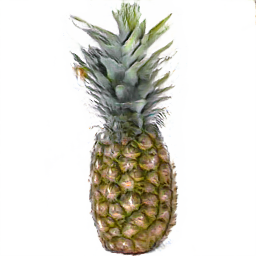}
  \end{subfigure}
  \begin{subfigure}[b]{\hrwidth\linewidth}
  \includegraphics[width=\linewidth]{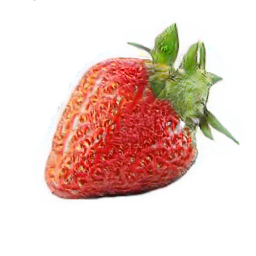}
  \end{subfigure}

  \begin{subfigure}[b]{\hrwidth\linewidth}
  \includegraphics[width=\linewidth]{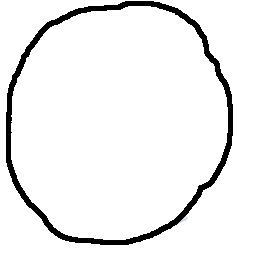}
  \end{subfigure}
  \begin{subfigure}[b]{\hrwidth\linewidth}
  \includegraphics[width=\linewidth]{img/hr/in/cupcake_146_real_B.png}
  \end{subfigure}
\begin{subfigure}[b]{\hrwidth\linewidth}
  \includegraphics[width=\linewidth]{img/hr/in/strawberry_146_real_B.png}
  \end{subfigure}
  \begin{subfigure}[b]{\hrwidth\linewidth}
  \includegraphics[width=\linewidth]{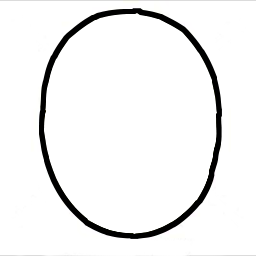}
  \end{subfigure}

  \begin{subfigure}[b]{\hrwidth\linewidth}
  \includegraphics[width=\linewidth]{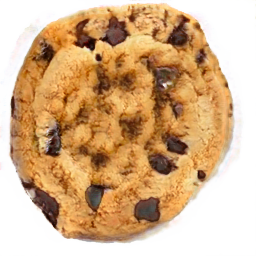}
  \end{subfigure}
  \begin{subfigure}[b]{\hrwidth\linewidth}
  \includegraphics[width=\linewidth]{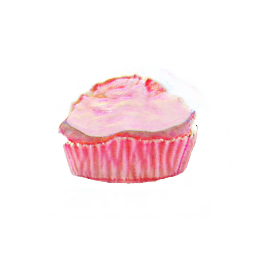}
  \end{subfigure}
\begin{subfigure}[b]{\hrwidth\linewidth}
  \includegraphics[width=\linewidth]{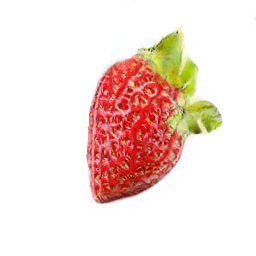}
  \end{subfigure}
  \begin{subfigure}[b]{\hrwidth\linewidth}
  \includegraphics[width=\linewidth]{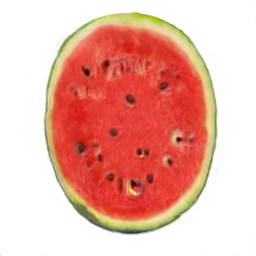}
  \end{subfigure}
  
\end{center}
\caption{In-domain $256\times 256$ results on the Scribble dataset.}
 \label{fig:in_res}
\end{figure*}

{\small
\bibliographystyle{ieee_fullname}
\bibliography{egbib}
}

\end{document}